\documentclass{article}

\usepackage{times}
\usepackage{graphicx} 

\usepackage{subcaption}
\usepackage{wrapfig}
\usepackage[ruled]{algorithm2e}
\usepackage{enumitem}

\usepackage{amsmath}
\usepackage{bbm}

\usepackage{amssymb}
\usepackage{dsfont}

\usepackage{hyperref}

\newcommand{\R}{\mathbb{R}}

\newcommand{\EE}{\mathbb{E}}

\newcommand{\sset}{\mathcal{S}}
\newcommand{\aset}{\mathcal{A}}

\newcommand{\trans}{\mathcal{P}}

\newtheorem{assumption}{Assumption}

\newenvironment{proc}[1][htb]
  {
   \begin{algorithm}[#1]
  }{\end{algorithm}}

\usepackage[final]{corl_2017} 

\title{Reverse Curriculum Generation \\for Reinforcement Learning}

\author{
   Carlos Florensa\\
   UC Berkeley \\ 
   \texttt{florensa@berkeley.edu} \\
   \And
   David Held\\
   UC Berkeley \\
   \texttt{davheld@berkeley.edu} \\
   \And
   Markus Wulfmeier\\
   Oxford Robotics Institute\\
   \texttt{markus@robots.ox.ac.uk}\\
   \And
   Michael R. Zhang\\
   UC Berkeley\\
   \texttt{michaelrzhang@berkeley.edu}\\
   \And
   Pieter Abbeel \\
   OpenAI \\
   UC Berkeley \\
  ICSI\\
   \texttt{pabbeel@berkeley.edu} \\
}

\begin{document}
\maketitle

\vspace{-0.8cm}
\begin{abstract}
Many relevant tasks require an agent to reach a certain state or to manipulate objects into a desired configuration. For example, we might want a robot to align and assemble a gear onto an axle or insert and turn a key in a lock.
These goal-oriented tasks present a considerable challenge for reinforcement learning, since their natural reward function is sparse and prohibitive amounts of exploration are required to reach the goal and receive some learning signal.
Past approaches tackle these problems by exploiting expert demonstrations or by manually designing a task-specific reward shaping function to guide the learning agent. Instead, we propose a method to learn these tasks without requiring any prior knowledge other than obtaining a single state in which the task is achieved.  The robot is trained in ``reverse," gradually learning to reach the goal from a set of start states increasingly far from the goal. Our method automatically generates a curriculum of start states that adapts to the agent's performance, leading to efficient training on goal-oriented tasks.  We demonstrate our approach on difficult simulated navigation and fine-grained manipulation problems, not solvable by state-of-the-art reinforcement learning methods.
\end{abstract}



\section{Introduction}

Reinforcement Learning (RL) is a powerful learning technique for training an agent to optimize a reward function.  Reinforcement learning has been demonstrated on complex tasks such as locomotion~\citep{schulman2015high}, Atari games~\cite{mnih2015human}, racing games~\cite{lillicrap2015continuous}, and robotic manipulation tasks~\cite{levine2016end}. 
However, there are many tasks for which it is hard to design a reward function such that it is both easy to maximize and yields the desired behavior once optimized. An ubiquitous example is a goal-oriented task; for such tasks, the natural reward function is usually sparse, giving a binary reward only when the task is completed \citep{held2017goal_generation}. This sparse reward can create difficulties for learning-based approaches \citep{duan2016benchmarking}; on the other hand, non-sparse reward functions for such tasks might lead to undesired behaviors \cite{popov2017ddpg_dexterous}.

For example, suppose we want a seven DOF robotic arm to learn how to align and assemble a gear onto an axle or place a ring onto a peg, as shown in Fig.~\ref{fig:ring_task}. The complex and precise motion required to align the ring at the top of the peg and then slide it to the bottom of the peg makes learning highly impractical if a binary reward is used.
On the other hand, using a reward function based on the distance between the center of the ring and the bottom of the peg leads to learning a policy that places the ring next to the peg, and the agent never learns that it needs to first lift the ring over the top of the peg and carefully insert it. 
Shaping the reward function~\citep{ng1999shaping} to efficiently guide the policy towards the desired solution often requires considerable human expert effort and experimentation to find the correct shaping function for each task. Another source of prior knowledge is the use of demonstrations, but it requires an expert intervention.

In our work, we avoid all reward engineering or use of demonstrations by exploiting two key insights. First, it is easier to reach the goal from states nearby the goal, or from states nearby where the agent already knows how to reach the goal. Second, applying random actions from one such state leads the agent to new feasible nearby states, from where it is not too much harder to reach the goal. This can be understood as requiring a minimum degree of reversibility, which is usually satisfied in many robotic manipulation tasks like assembly and manufacturing.

We take advantage of these insights to develop a ``reverse learning" approach for solving such difficult manipulation tasks.  The robot is first trained to reach the goal from start states nearby a given goal state.  Then, leveraging that knowledge, the robot is trained to solve the task from increasingly distant start states. All start states are automatically generated by executing a short random walk from the previous start states that got some reward but still require more training. This method of learning in reverse, or growing outwards from the goal, is inspired by dynamic programming methods like value iteration, where the solutions to easier sub-problems are used to compute the solution to harder problems.

In this paper, we present an efficient and principled framework for performing such ``reverse learning."  Our method automatically generates a curriculum of initial positions from which to learn to achieve the task.
This curriculum constantly adapts to the learning agent by observing its performance at each step of the training process.  Our method requires no prior knowledge of the task other than providing a single state that achieves the task (i.e.~is at the goal). The contributions of this paper include:
\vspace{-0.3cm}
\begin{itemize}
\item Formalizing a novel problem definition of finding the optimal start-state distribution at every training step to maximize the overall learning speed.
\item A novel and practical approach for sampling a start state distribution that varies over the course of training, leading to an automatic curriculum of start state distributions.
\item Empirical experiments showing that our approach solves difficult tasks like navigation or fine-grained robotic manipulation, not solvable by state-of-the-art learning methods.
\end{itemize}


\section{Related Work}

Curriculum-based approaches with manually designed schedules have been explored in supervised learning \citep{bengio2009curriculum, zaremba2014execute, bengio2015scheduled, graves2017curriculum} to split particularly complex tasks into smaller, easier-to-solve sub-problems.
One particular type of curriculum learning explicitly enables the learner to reject examples which it currently considers too hard \citep{kumar2010self, jiang2015self}. 
This type of adaptive curriculum has mainly been applied to supervised tasks, and most practical curriculum approaches in RL rely on pre-specified task sequences \citep{asada1996purposive, karpathy2012curriculum}. Some very general frameworks have been proposed to generate increasingly hard problems \citep{schmidhuber2013powerplay,srivastava2012powerplay_experiments}, although challenges remain to apply the idea to difficult robotics tasks. A similar line of work uses intrinsic motivation based on learning progress to obtain ``developmental trajectories" that focus on increasingly difficult tasks \cite{baranes2013goal}. Nevertheless, their method requires iteratively partitioning the full task space, which strongly limits the application to fine-grain manipulation tasks like the ones presented in our work (see detailed analysis on easier tasks in \cite{held2017goal_generation}).

More recent work in using a curriculum for RL assumes that baseline performances for several tasks are given, and it uses them to gauge which tasks are the hardest (furthest behind the baseline) and require more training \citep{sharma2017multi-task}. However, this framework can only handle finite sets of tasks and requires each task to be learnable on its own. On the other hand, our method trains a policy that generalizes to a set of continuously parameterized tasks, and it is shown to perform well even under sparse rewards by not allocating training effort to tasks that are too hard for the current performance of the agent.

Closer to our method of adaptively generating the tasks to train on, an interesting asymmetric self-play strategy has recently been proposed~\citep{sukhbaatar2017asymmetric}. Contrary to our approach, which aims to generate and train on all tasks that are at the appropriate level of difficulty, the asymmetric component of their method can lead to biased exploration concentrating on only a subset of the tasks that are at the appropriate level of difficulty, as the authors and our own experiments suggests. This problem and their time-oriented metric of hardness may lead to poor performance in continuous state-action spaces, which are typical in robotics.
Furthermore, their approach is designed as an exploration bonus for a single target task; in contrast, we define a new problem of efficiently optimizing a policy across a range of start states, which is considered relevant to improve generalization \citep{rajeswaran2017simplicity}.

Our approach can be understood as sequentially composing locally stabilizing controllers by growing a tree of stabilized trajectories backwards from the goal state, similar to work done by \citet{tedrake2010lqr}. This can be viewed as a ``funnel'' which takes start states to the goal state via a series of locally valid policies \citep{burridge1999sequential}. Unlike these methods, our approach does not require any dynamic model of the system. An RL counterpart, closer to our approach, is the work by \citet{bagnell2003search}, where a policy search algorithm in the spirit of traditional dynamic programming methods is proposed to learn a non-stationary policy: they learn what should be done in the last time-step and then ``back it up" to learn the previous time-step and so on. Nevertheless, they require the stronger assumption of having access to baseline distributions that approximate the optimal state-distribution at every time-step.

The idea of directly influencing the start state distribution to accelerate learning in a Markov Decision Process (MDP) has drawn attention in the past. 
\citet{kakade2002approximately} studied the idea of exploiting the access to a `generative model' \citep{kearns2002sampling} that allows training the policy on a fixed `restart distribution' different from the one originally specified by the MDP. If properly chosen, this is proven to improve the policy training and final performance on the original start state distribution. Nevertheless, no practical procedure is given to choose this new distribution (only suggesting to use a more uniform distribution over states, which is what our baseline does), and they don't consider adapting the start state distribution during training, as we do. Other researchers have proposed to use expert demonstrations to improve learning of model-free RL algorithms, either by modifying the start state distribution to be uniform among states visited by the provided trajectories \cite{popov2017ddpg_dexterous}, or biasing the exploration towards relevant regions \cite{subramanian2016exploration_from_demos}. Our method works without any expert demonstrations, so we do not compare against these lines of research.


\section{Problem Definition}
\label{sec:problem_definition}
We consider the general problem of learning a policy that leads a system into a specified goal-space, from any start state sampled from a given distribution. In this section we first briefly introduce the general reinforcement learning framework and then we formally define our problem statement.

\subsection{Preliminaries}
\label{sec:preliminaries}
We define a discrete-time finite-horizon Markov decision process (MDP) by a tuple $M = (\sset, \aset, \trans, r, \rho_0, T)$, in which $\sset$ is a state set, $\aset$ an action set, $\trans: \sset \times \aset \times \sset \rightarrow \mathbb{R}_{+}$ is a transition probability distribution, $r: \sset \times \aset \rightarrow \R$ is a bounded reward function, $\rho_0: \sset \to \mathbb{R}_+$ is a start state distribution, and $T$ is the horizon. Our aim is to learn a stochastic policy $\pi_{\theta}: \sset \times \aset \to \mathbb{R}_+$ parametrized by $\theta$ that maximizes the expected return, $ \eta_{\rho_0}(\pi_\theta) = \EE_{s_0\sim\rho_0} R(\pi, s_0)$. We denote by $ R(\pi, s_0) := \EE_{\tau|s_0}[ \sum_{t=0}^T r(s_t, a_t) ]$ the expected cumulative reward starting when starting from a $s_0\sim \rho_0$, where $\tau = (s_0, a_0, , \ldots, a_{T-1}, s_T)$ denotes a whole trajectory, with $a_t \sim \pi_\theta(a_t|s_t)$, and $s_{t+1} \sim \trans(s_{t+1} | s_t, a_t)$. Policy search methods iteratively collect trajectories on-policy (i.e.~sampling from the above distributions $\rho_0$, $\pi_{\theta}$, $\trans$) and use them to improve the current policy~\cite{bertsekas1995dynamic,peters2008reinforcement,szita2006learning}. 

In our work we propose to instead use a different start-state distribution $\rho_i$ at every training iteration $i$ to maximize the learning rate. Learning progress is still evaluated based on the original distribution $\rho_0$. Convergence of $\rho_i$ to $\rho_0$ is desirable but not required as an optimal policy $\pi^{\star}_i$ under a start distribution $\rho_i$ is also optimal under any other $\rho_0$, as long as their support coincide. In the case of approximately optimal policies under $\rho_i$, bounds on the performance under $\rho_0$ can be derived \cite{kakade2002approximately}.

\subsection{Goal-oriented tasks}
\label{sec:task}
We consider the general problem of reaching a certain goal space $S^g\subset \sset$ from any start state in $S^0 \subset \mathcal{S}$. This simple, high-level description can be translated into an MDP without further domain knowledge by using a binary reward function $r(s_t) = \mathds{1}\big\{s_t\in S^g\big\}$ and a uniform distribution over the start states $\rho_0 = \textrm{Unif}(S^0)$. We terminate the episode when the goal is reached. This implies that the return $R(\pi, s_0)$ associated with every start state $s_0$ is the probability of reaching the goal at some time-step $t\in\{0\dots T\}$. 
\begin{align}
\label{eq:expected_return}
R(\pi, s_0) 
= \EE_{\pi(\cdot | s_t)} \mathds{1}\Big\{\bigcup_{t=0}^T s_{t} \in S^g | s_0\Big\}
= \mathds{P}\Big(\bigcup_{t=0}^T s_{t} \in S^g ~\Big|~ \pi, s_0\Big)
\end{align}
As advocated by \citet{rajeswaran2017simplicity}, it is important to be able to train an agent to achieve the goal from a large set of start states $S^0$.  An agent trained in this way would be much more robust than an agent that is trained from just a single start state, as it could recover from undesired deviations from the intended trajectory. Therefore, we choose the set of start states $S^0$ to be all the feasible points in a wide area around the goal. 
On the other hand, the goal space $S^g$ for our robotics fine-grained manipulation tasks is defined to be a small set of states around the desired configuration (e.g. key in the key-hole, or ring at the bottom of the peg, as described in Sec.~\ref{sec:experiments}).

As discussed above, the sparsity of this reward function makes learning extremely difficult for RL algorithms ~\citep{duan2016benchmarking, osband2014generalization, whitehead1991complexity}, and approaches like reward shaping~\citep{ng1999shaping} are difficult and time-consuming to engineer for each task. In the following subsection we introduce three assumptions, and the rest of the paper describes how we can leverage these assumptions to efficiently learn to achieve complex goal-oriented tasks directly from sparse reward functions.

\subsection{Assumptions for reverse curriculum generation}
\label{sec:assumptions}

In this work we study how to exploit three assumptions that hold true in a wide range of practical learning problems (especially if learned in simulation):
\begin{assumption}
    \label{assumption:reset}
    We can arbitrarily reset the agent into any start state $s_0 \in \mathcal{S}$ at the beginning of all trajectories.
\end{assumption}
\vspace{-0.4cm}
\begin{assumption}
    \label{assumption:goal}
    At least one state $s^g$ is provided such that $s^g\in S^g$.
\end{assumption}
\vspace{-0.4cm}
\begin{assumption}
    \label{assumption:communication}
    The Markov Chain induced by taking uniformly sampled random actions has a communicating class\footnote{A \textit{communicating class} is a maximal set of states $C$ such that every pair of states in $C$ communicates with each other. Two states \textit{communicate} if there is a non-zero probability of reaching one from the other.} including all start states $S^0$ and the given goal state $s^g$.  
\end{assumption}

The first assumption has been considered previously (e.g. access to a generative model in \citet{kearns2002sampling}) and is deemed to be a considerably weaker assumption than having access to the full transition model of the MDP. \citet{kakade2002approximately} proved that Assumption \ref{assumption:reset} can be used to improve the learning in MDPs that require large exploration. Nevertheless, they do not propose a concrete procedure to choose a distribution $\rho$ from which to sample the start states in order to maximally improve on the objective in Eq.~\eqref{eq:expected_return}.
In our case, combining Assumption \ref{assumption:reset} with Assumption \ref{assumption:goal}, we are able to reset the state to $s^g$, which is critical in our method to initialize the start state distribution to concentrate around the goal space at the beginning of learning. For Assumption \ref{assumption:goal}, note that we only assume access to one state $s^g$ in the goal region; we do not require a description of the full region nor trajectories leading to it. Finally, Assumption \ref{assumption:communication} ensures that the goal can be reached from any of the relevant start states, and that those start states can also be reached from the goal; this assumption is satisfied by many robotic problems of interest, as long as there are no major irreversibilities in the system. In the next sections we detail our automatic curriculum generation method based on continuously adapting the start state distribution to the current performance of the policy.  We demonstrate the value of this method for challenging robotic manipulation tasks.


\section{Methodology}
\label{sec:methodology}

In a wide range of goal-oriented RL problems, reaching the goal from an overwhelming majority of start states in $S^0$ requires a prohibitive amount of on-policy or undirected exploration. On the other hand, it is usually easy for the learning agent (i.e. our current policy $\pi_i$) to reach the goal $S^g$ from states nearby a goal state $s^g\in S^g$. Therefore, learning from these states will be fast because the agent will perceive a strong signal, even under the indicator reward introduced in Section \ref{sec:task}. 
Once the agent knows how to reach the goal from these nearby states, it can train from even further states and bootstrap its already acquired knowledge. This reverse expansion is inspired by classical RL methods like Value Iteration or Policy Iteration \citep{sutton1998reinforcement}, although in our case we do not assume knowledge of the transition model and our environments have high-dimensional continuous action and state spaces. In the following subsections we propose a method that leverages the assumptions from the previous section and the idea of reverse expansion to automatically adapt the start state distribution, generating a curriculum of start state distributions that can be used to tackle problems unsolvable by standard RL methods. 

\subsection{Policy Optimization with modified start state distribution}
\label{sec:pol_opt}
Policy gradient strategies are well suited for robotic tasks with continuous and high dimensional action-spaces ~\citep{deisenroth2013survey}. 
Nevertheless, applying them directly on the original MDP does poorly in tasks with sparse rewards and long horizons like our challenging manipulation tasks. If the goal is not reached from the start states in $S^0$, no reward is received, and the policy cannot improve. Therefore, we propose to adapt the distribution $\rho_i$ from where start states $s_0$ are sampled to train policy $\pi_i$. 

Analogously to \citet{held2017goal_generation}, we postulate that in goal-oriented environments, a strong learning signal is obtained when training on start states $s_0\sim\rho_i$ from where the agent reaches the goal sometimes, but not always. We call these start states ``good starts". More formally, at training iteration $i$, we would like to sample from $\rho_i = \textrm{Unif}(S^0_i)$ where $S^0_i=\{s_0:R_{\min} < R(\pi_i, s_0) < R_{\max} \}$. The hyper-parameters $R_{\min}$ and $R_{\max}$ are easy to tune due to their interpretation as bounds on the probability of success, derived from Eq.~\eqref{eq:expected_return}. Unfortunately, sampling uniformly from $S^0_i$ is intractable. Nevertheless, at least at the beginning of training, states nearby a goal state $s^g$ are more likely to be in $S^0_i$. Then, after some iterations of training on these start states, some will be completely mastered (i.e. $R(\pi_{i+1}, s_0) > R_{\rm max}$ and $s_0$ is no longer in $S^0_{i+1}$), but others will still need more training. To find more ``good starts", we follow the same reasoning: the states nearby these remaining $s\in S^0_{i+1}$ are likely to also be in $S^0_{i+1}$. In the rest of the section we describe an effective way of sampling feasible nearby states and we layout the full algorithm. 



\subsection{Sampling ``nearby" feasible states} 
\label{sec:nearby_sampling}

For robotic manipulation tasks with complex contacts and constraints, applying noise in state-space $s'=s+\epsilon,~\epsilon\sim\mathcal{N}$ may yield many infeasible states $s'$. For example, even small random perturbations of the joint angles of a seven degree-of-freedom arm generate large modifications to the end-effector position, potentially placing it in an infeasible state that intersects with surrounding objects. For this reason, the concept of ``nearby" states might be unrelated to the Euclidean distance $\|s'-s\|^2$ between these states. Instead, we have to understand proximity in terms of how likely it is to reach one state from the other by taking actions in the MDP.

Therefore, we choose to generate new states $s'$ from a certain seed state $s$ by applying noise in action space. 
This means we exploit Assumption \ref{assumption:reset} to reset the system to state $s$, and from there we execute short ``Brownian motion" rollouts of horizon $T_B$ taking actions $a_{t+1} = \epsilon_{t}$ with $\epsilon_t\sim\mathcal{N}(0, \Sigma)$. This method of generating ``nearby" states is detailed in Procedure \ref{alg:brownian}. The total sampled states $M$ should be large enough such that the $N_{\rm new}$ desired states $starts_{\rm new}$, obtained by subsampling, extend in all directions around the input states $starts$.
All states visited during the rollouts are guaranteed to be feasible and can then be used as start states to keep training the policy.

\subsection{Detailed Algorithm}
\label{sec:full_algo}

\begin{figure}
\begin{minipage}{0.60\textwidth}
\vspace{-0.7cm}
\begin{algorithm}[H]
\SetAlgoLined
 \SetKwInOut{Input}{Input}
 \SetKwInOut{Output}{Output}
 \Input{$\pi_0$, $s^g$, $\rho_0$, $N_{\rm new}$, $N_{\rm old}$, $R_{\min}$, $R_{\max}$, $Iter$}
 \Output{Policy $\pi_N$}
 
 $starts_{\rm old} \leftarrow [s^g]$\;
 
 $starts,~ rews \leftarrow [s^g],~ [1]$\;
 
 \For{$i \leftarrow \ 1$ \KwTo $Iter$}{
 
  $starts \leftarrow \texttt{SampleNearby}(starts,~N_{\rm new})$\;
  
  $starts\text{.append}{\rm[}\text{sample}(starts_{\rm old}, N_{\rm old})$]\;
  
  $\rho_i \leftarrow \text{Unif}(starts)$\;
    
  $\pi_{i},\ rews \leftarrow \texttt{train\_pol}(\rho_i,~ \pi_{i-1})$\;
  
  $starts \leftarrow \texttt{select}(starts,~ rews, R_{\min}, R_{\max})$\;
  
  $starts_{\rm old}\text{.append[}starts$]\;
  
  
 }
 \caption{Policy Training}
 \label{alg:overall}
\end{algorithm}
\vspace{-0.5cm}
\end{minipage}
\begin{minipage}{0.40\textwidth}
\vspace{-0.7cm}
\begin{proc}[H]
\SetAlgoLined
 \SetKwInOut{Input}{Input}
 \SetKwInOut{Output}{Output}
 \Input{$starts$, $N_{\rm new}$, $\Sigma$, $T_B$, $M$}
 \Output{$starts_{\rm new}$}
 
 \While{{\rm len}($starts$) $< M$}{
 
      $ s_0 \sim {\rm Unif}(starts)$\;
      
      \For{$t \leftarrow 1$ \KwTo $T_B$}{
            $a_{t} = \epsilon_{t}$,~ $\epsilon_t\sim\mathcal{N}(0, \Sigma)$\;
            $s_{t} \sim \mathcal{P}(s_{t}| s_{t-1}, a_{t}) $\;
            $starts$.append($s_{t}$)\;
      }
 }
 $starts_{\rm new} \leftarrow \text{sample($starts, N_{\rm new}$)}$
 \caption{\texttt{SampleNearby}}
 \label{alg:brownian}
\end{proc}
\vspace{-0.5cm}
\end{minipage}
\end{figure}

Our generic algorithm is detailed in Algorithm \ref{alg:overall}. We first initialize the policy with $\pi_0$ and the ``good start" states list $starts$ with the given goal state $s^g$. Then we perform $Iter$ training iterations of our RL algorithm of choice $\texttt{train\_pol}$. In our case we perform 5 iterations of Trust Region Policy Optimization (TRPO)~\citep{schulman2015trust} but any on-policy method could be used. 
At every iteration, we set the start state distribution $\rho_i$ to be uniform over a list of start states obtained by sampling $N_{\rm new}$ start states from nearby the ones in our ``good starts" list $starts$ (see $\texttt{SampleNearby}$ in previous section), and $N_{\rm old}$ start states from our replay buffer of previous ``good starts" $starts_{\rm old}$. As already shown by \citet{held2017goal_generation}, the replay buffer is an important feature to avoid catastrophic forgetting.
Technically, to check which of the states $s_0\in starts$ are in $S^0_i$ (i.e.~the ``good starts") we should execute some trajectories from each of those states to estimate the expected returns $R(s_0, \pi_{i-1})$, but this considerably increases the sample complexity. Instead, we use the trajectories collected by $\texttt{train\_pol}$ to estimate $R(\pi_{i-1}, s_0)$ and save it in the list $rews$. These are used to $\texttt{select}$ the ``good" start states for the next iteration - picking the ones with $R_{\min}\leq R(\pi_{i-1}, s_0) \leq R_{\max}$. We found this heuristic to give a good enough estimate and not drastically decrease learning performance of the overall algorithm.

Our method keeps expanding the region of the state-space from which the policy can reach the goal reliably. It samples more heavily nearby the start states that need more training to be mastered and avoiding start states that are yet too far to receive any reward under the current policy. Then, thanks to Assumption \ref{assumption:communication}, the Brownian motion that is used to generate further and further start states will eventually reach all start states in $S^0$, and therefore our method improves the metric $\eta_{\rho_0}$ defined in Sec.~\ref{sec:preliminaries} (see also Sec.~\ref{sec:metric} for details on how we evaluate our progress on this metric).


\section{Experimental Results}
\label{sec:experiments}

We investigate the following questions in our experiments:
\vspace{-0.2cm}
\begin{itemize}[leftmargin=2em, noitemsep]
    \item Does the performance of the policy on the target start state distribution $\rho_0$ improve by training on distributions $\rho_i$ growing from the goal?
    \item Does focusing the training on ``good starts" speed up learning?
    \item Is Brownian motion a good way to generate ``good starts" from previous ``good starts"? 
\end{itemize}
\vspace{-0.2cm}
We use the below task settings to explore these questions. All are implemented in MuJoCo \cite{todorov2012mujoco} and the hyperparameters used in our experiments are described in Appendix \ref{sec:hyper}.

\begin{figure}[ht]
\captionsetup[subfigure]{justification=centering}
    \centering
      \begin{subfigure}{0.24\linewidth}
        \includegraphics[width=\linewidth, trim={0cm, 0cm, 0cm, 0cm}, clip]{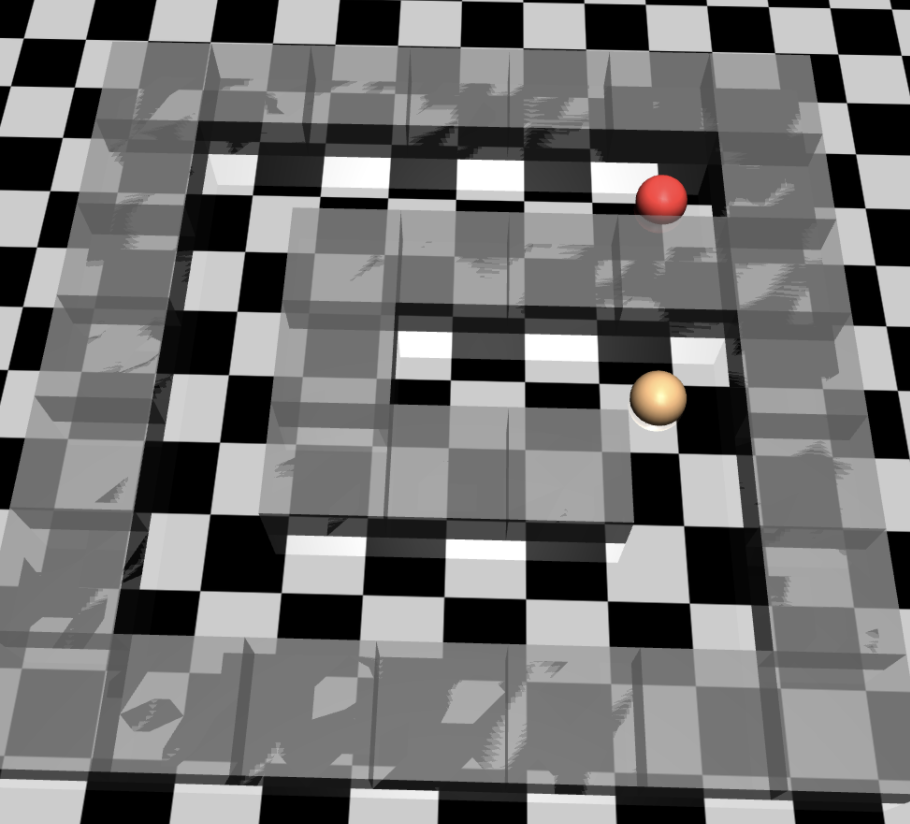}
          \caption{Point-mass maze task}
          \label{fig:maze11}
      \end{subfigure}
      \begin{subfigure}{0.24\linewidth}
        \includegraphics[width=\linewidth, trim={0cm, 0cm, 0cm, 0cm}, clip]{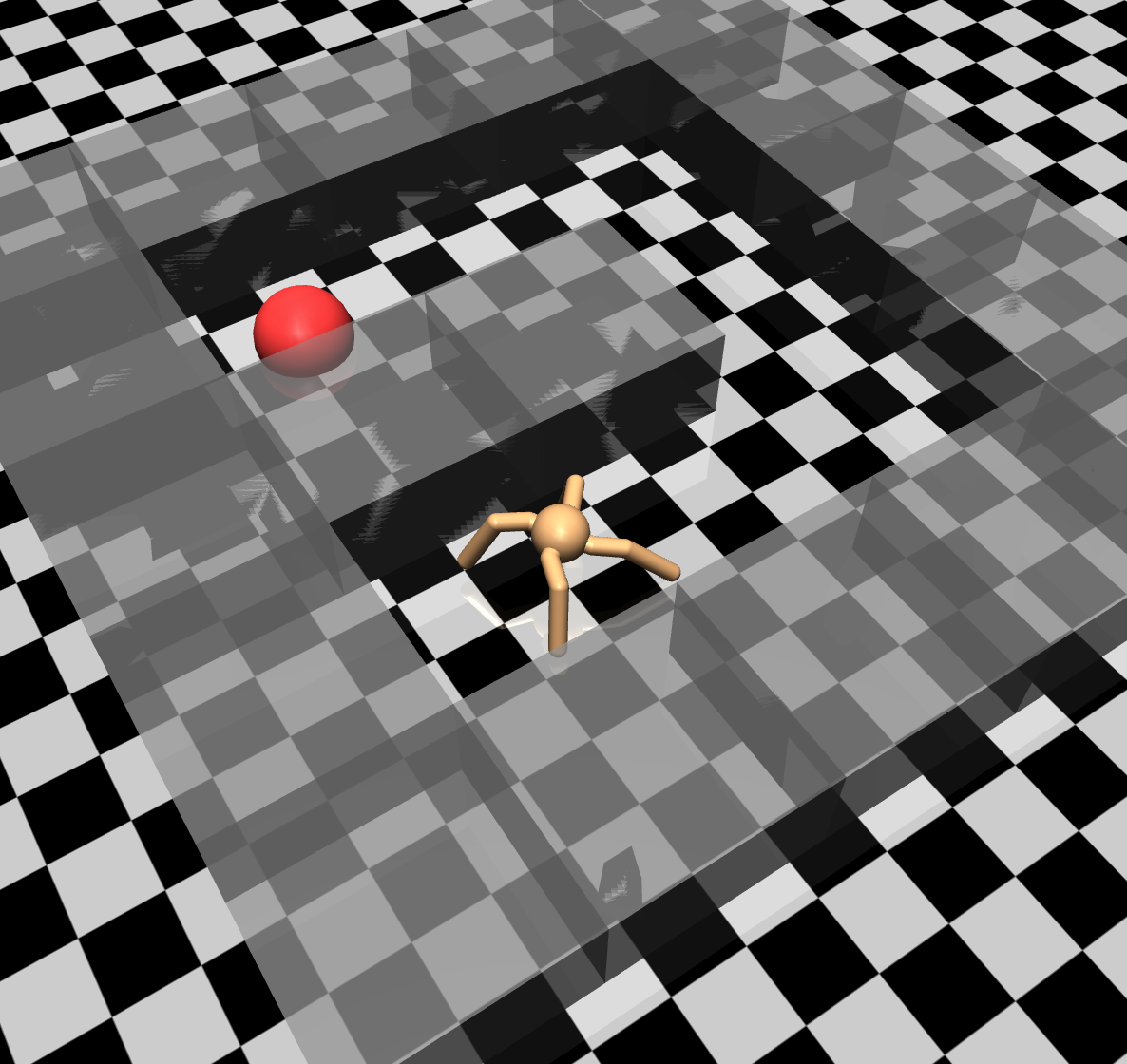}
          \caption{Ant maze task}
          \label{fig:ant_maze_task}
      \end{subfigure}
      \begin{subfigure}{0.24\linewidth}
        \includegraphics[width=\linewidth, trim={0cm, 0cm, 0cm, 0cm}, clip]{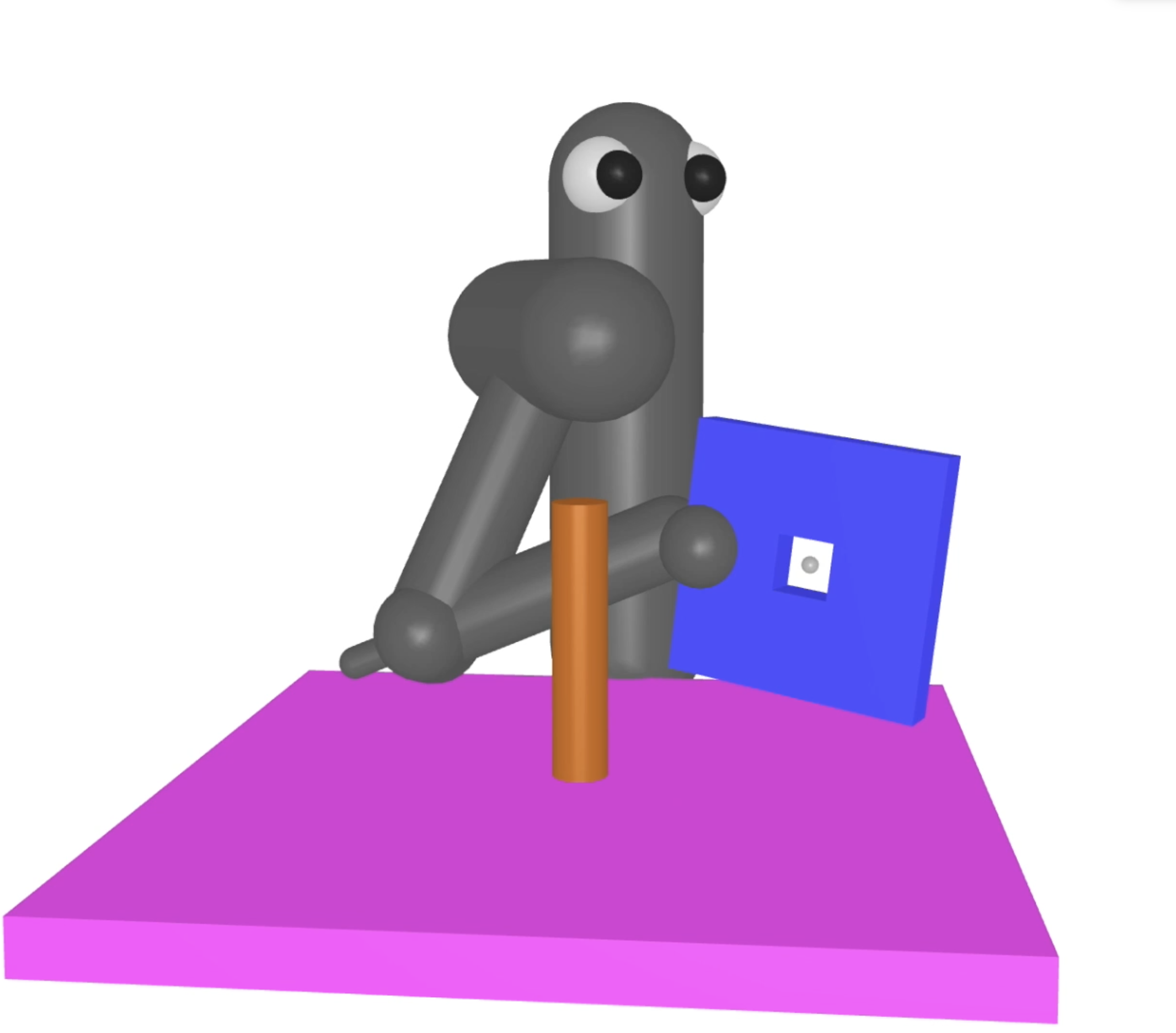}
          \caption{Ring on Peg task}
          \label{fig:ring_task}
      \end{subfigure}
      \begin{subfigure}{0.24\linewidth}
        \includegraphics[width=\linewidth, trim={0cm, 0cm, 0cm, 0cm}, clip]{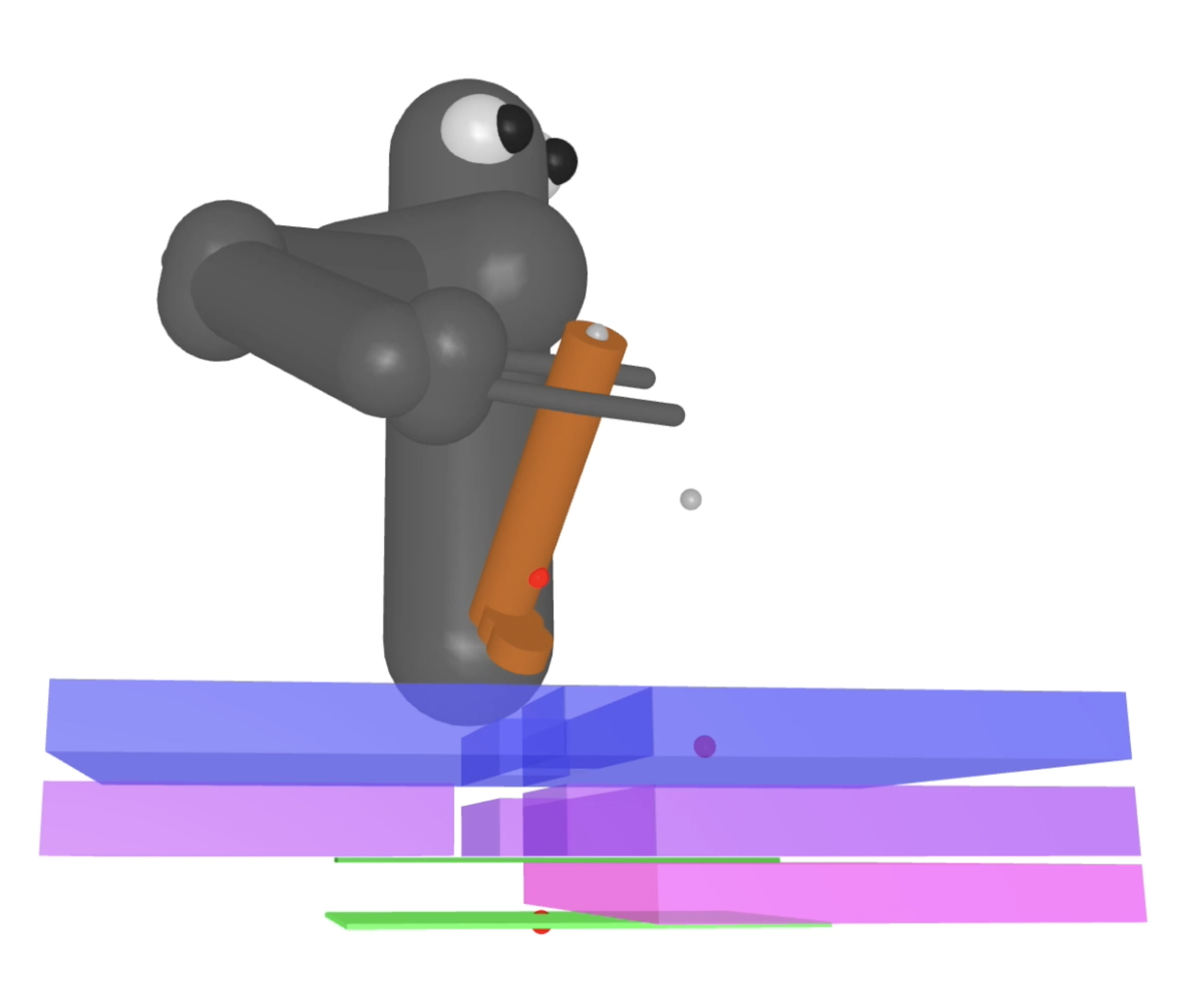}
          \caption{Key insertion task}
          \label{fig:key_task}
      \end{subfigure}
\vspace{-4pt}
\caption{Task images. Source code and videos of the performance obtained by our algorithm are available here: \url{http://bit.ly/reversecurriculum}} 
\label{fig:tasks}
\vspace{-0.2cm}
\end{figure}

\textbf{Point-mass maze:} (Fig.~\ref{fig:maze11}) A point-mass agent (orange) must navigate within 30cm of the goal position $(4m, 4m)$ at the end of a G-shaped maze (red). The target start state distribution from which we seek to reach the goal is uniform over all feasible $(x,y)$ positions in the maze. 

\textbf{Ant maze:} (Fig.~\ref{fig:ant_maze_task}) A  quadruped robot (orange) must navigate its Center of Mass to within 50cm of the goal position $(0m, 4m)$ at the end of a U-shaped maze (red). The target start state distribution from which we seek to reach the goal is uniform over all feasible ant positions inside the maze.

\textbf{Ring on Peg:} (Fig.~\ref{fig:ring_task}) A 7 DOF robot must learn to place a ``ring" (actually a square disk with a hole in the middle) on top of a tight-fitting round peg.  The task is complete when the ring is within 3 cm of the bottom of the 15 cm tall peg.  The target start state distribution from which we seek to reach the goal is uniform over all feasible joint positions for which the center of the ring is within 40 cm of the bottom of the peg.

\textbf{Key insertion:} (Fig.~\ref{fig:key_task}) A 7 DOF robot must learn to insert a key into a key-hole. The task is completed when the distance between three reference points at the extremities of the key and its corresponding targets is below 3cm. In order to reach the target, the robot must first insert the key at a specific orientation, then rotate it 90 degrees clockwise, push forward, then rotate 90 degrees counterclockwise. The target start state distribution from which we seek to reach the goal is uniform over all feasible joint positions such that the tip of the key is within 40 cm of key-hole.

\subsection{Effect of start state distribution}

\begin{figure}[ht]
\captionsetup[subfigure]{justification=centering}
    \centering
      \begin{subfigure}{0.4\linewidth}
        \includegraphics[width=\linewidth, trim={0cm, 0cm, 0cm, 0cm}, clip]{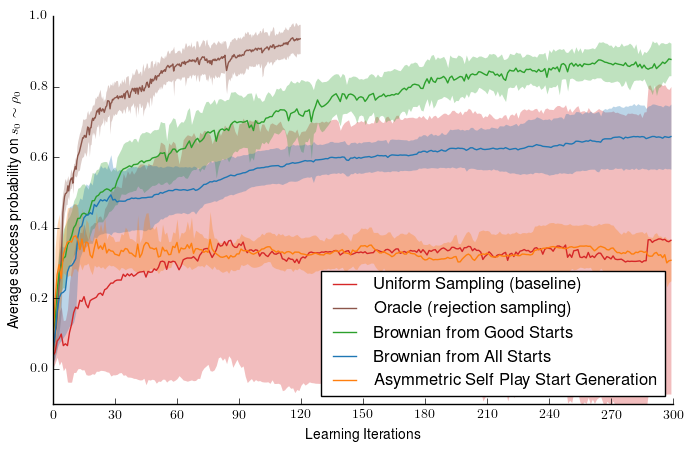}
        \caption{Point-mass Maze task}
          \label{fig:learning_point_maze_task}
      \end{subfigure}
      \begin{subfigure}{0.4\linewidth}
        \includegraphics[width=\linewidth, trim={0cm, 0cm, 0cm, 0cm}, clip]{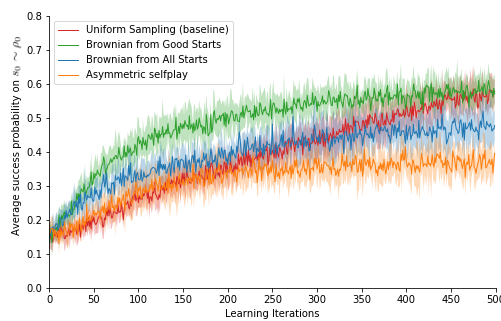}
          \caption{Ant Maze task}
          \label{fig:learning_ant_maze_task}
      \end{subfigure}
      \begin{subfigure}{0.4\linewidth}
        \includegraphics[width=\linewidth, trim={0cm, 0cm, 0cm, 0cm}, clip]{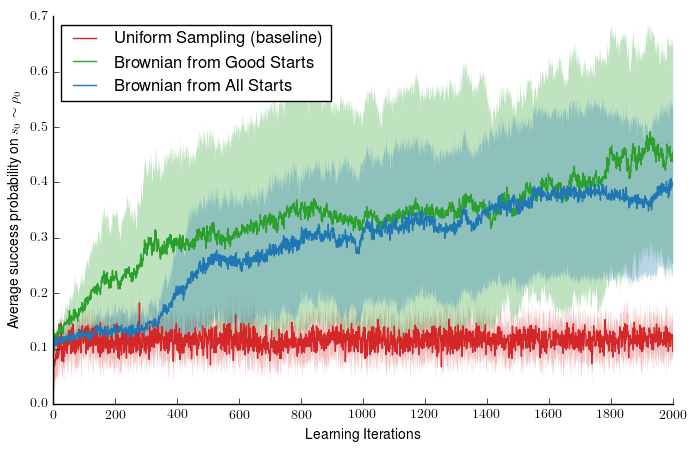}
          \caption{Ring on Peg task}
          \label{fig:learning_ring_task}
      \end{subfigure}
      \begin{subfigure}{0.4\linewidth}
        \includegraphics[width=\linewidth, trim={0cm, 0cm, 0cm, 0cm}, clip]{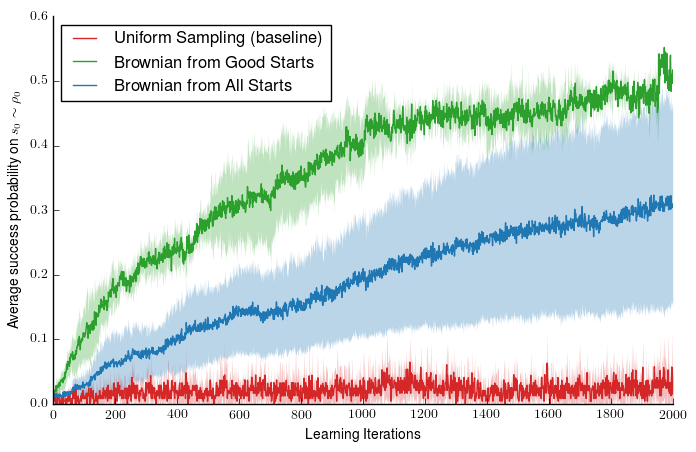}
          \caption{Key insertion task}
          \label{fig:learning_key_task}
      \end{subfigure}
\caption{Learning curves for goal-oriented tasks (mean and variance over 5 random seeds).} 
\label{fig:learning_manipulation_tasks}
\vspace{-0.2cm}
\end{figure}

In Figure \ref{fig:learning_manipulation_tasks}, the \textit{Uniform Sampling (baseline)} red curves show the average return of policies learned with TRPO without modifying the start state distribution. The green and blue curves correspond to our method and an ablation, both exploiting the idea of modifying the start state distribution at every learning iteration. These approaches perform consistently better across the board. In the case of the point-mass maze navigation task in Fig.~\ref{fig:learning_point_maze_task}, we observe that \textit{Uniform Sampling} has a very high variance because some policies only learn how to perform well from one side of the goal (see Appendix \ref{sec:failure_uniform} for a thorough analysis). The Ant-maze experiments in Fig.~\ref{fig:learning_ant_maze_task} also show a considerable slow-down of the learning speed when using plain TRPO, although the effect is less drastic as the start state distribution $\rho_0$ is over a smaller space. 

In the more complex manipulation tasks shown in Fig.~\ref{fig:learning_ring_task}-\ref{fig:learning_key_task}, we see that the probability of reaching the goal with \textit{Uniform Sampling} is around 10\% for the ring task and 2\% for the key task. These success probabilities correspond to reliably reaching the goal only from very nearby positions: when the ring is already on the peg or when the key is initialized very close to the final position. None of the learned policies trained on the original $\rho_0$ learn to reach the goal from more distant start states. On the other hand, our methods do succeed at reaching the goal from a wide range of far away start states. The underlying RL training algorithm and the evaluation metric are the same. We conclude that training on a different start state distribution $\rho_i$ can improve training or even allow learning at all.

\subsection{Effect of ``good starts"}
\label{sec:results_good_starts}
In Figure~\ref{fig:learning_manipulation_tasks} we see how applying our Algorithm \ref{alg:overall} to modify the start state distribution considerably improves learning  (\textit{Brownian on Good Starts}, in green) and final performance on the original MDP. Two elements are involved in this improvement: first, the backwards expansion from the goal, and second, the concentration of training efforts on ``good starts". To test the relevance of this second element, we ablate our method by running our \texttt{SampleNearby} Procedure \ref{alg:brownian} on all states from which the policy was trained in the previous iteration. In other words, the \texttt{select} function in Algorithm \ref{alg:overall} is replaced by the identity, returning all \textit{starts}
independently of the rewards \textit{rews} they obtained during the last training iteration. 
The resulting algorithm performance is shown as the \textit{Brownian from All Starts} blue curve in Figures \ref{fig:learning_manipulation_tasks}. As expected, this method is still better than not modifying the start state distribution but has a slower learning rate than running \texttt{SampleNearby} around the estimated good starts.

Now we evaluate an upper bound of the benefit provided by our idea of sampling ``good starts". As mentioned in Sec.~\ref{sec:pol_opt}, we would ideally like to sample start states from $\rho_i = \textrm{Unif}(S^0_i)$, but it is intractable.  Instead, we evaluate states in $S^0_{i-1}$, and we use Brownian motion to find nearby states, to approximate $S^0_i$. We can evaluate how much this approximation hinders learning 
by exhaustively sampling states in the lower dimensional point-mass maze task. To do so, at every iteration we can sample states $s_0$ uniformly from the state-space $\sset$, empirically estimate their return $R(s_0,\pi_i)$, and reject the ones that are not in the set $S^0_i=\{s_0:R_{\min} < R(\pi_i, s_0) < R_{\max} \}$.  This exhaustive sampling method is orders of magnitude more expensive in terms of sample complexity, so it would not be of practical use. In particular, we can only run it in the easier point-mass maze task. Its performance is shown in the brown curve of Fig.~\ref{fig:learning_point_maze_task}, called ``Oracle (rejection sampling)";  training on states sampled in such a manner further improves the learning rate and final performance.  Thus we can see that our approximation of using states in $S^0_{i-1}$ to find states in $S^0_i$ leads to some loss in performance, at the benefit of a greatly reduced computation time.

Finally, we compare to another way of generating start states based on the asymmetric self-play method of~\citet{sukhbaatar2017intrinsic}. The basic idea is to train another policy, ``Alice", that proposes start states to the learning policy, ``Bob". As can be seen, this method performs very poorly in the point-mass maze task, and our investigation shows that ``Alice" often gets stuck in a local optimum, leading to poor start states suggestions for ``Bob". In the original paper, the method was demonstrated only on discrete action spaces, in which a multi-modal distribution for Alice can be maintained; even in such settings, the authors observed that Alice can easily get stuck in local optima.  This problem is exacerbated when moving to continuous action spaces defined by a unimodal Gaussian distribution. See a detailed analysis of these failure modes in Appendix \ref{sec:failure_asym}.

\subsection{Brownian motion to generate good starts ``nearby" good starts}
\begin{figure}[ht]
\captionsetup[subfigure]{justification=centering}
    \centering
      \begin{subfigure}{0.45\linewidth}
        \includegraphics[width=\linewidth, trim={0cm, 0cm, 0cm, 1cm}, clip]{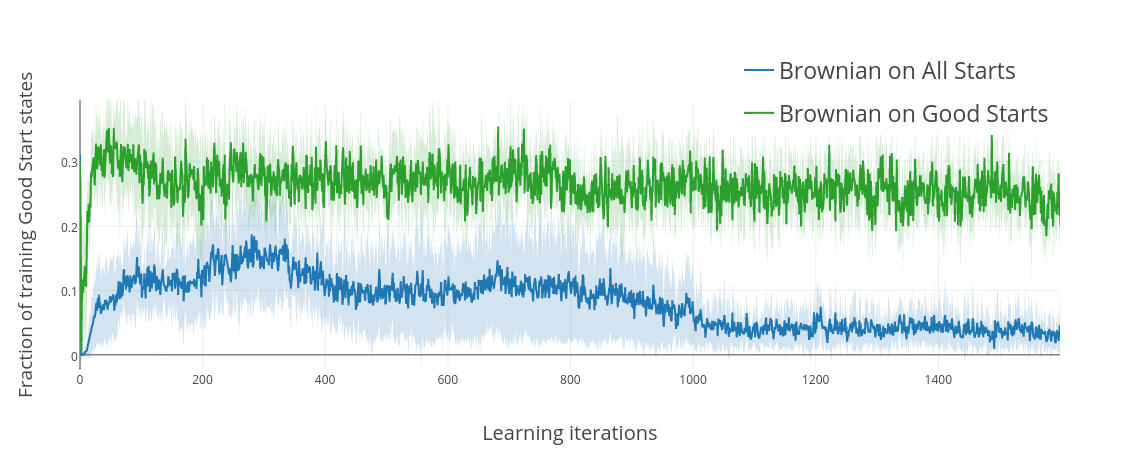}
          \caption{Ring on Peg task}
          \label{fig:good_starts_disc}
      \end{subfigure}
      \begin{subfigure}{0.45\linewidth}
        \includegraphics[width=\linewidth, trim={0cm, 0cm, 0cm, 1cm}, clip]{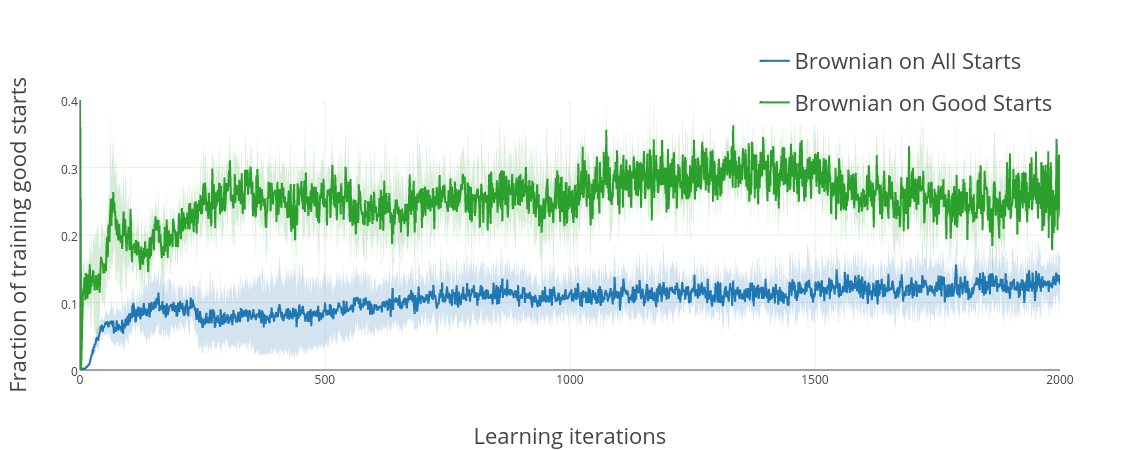}
          \caption{Key insertion task}
          \label{fig:good_starts_key}
      \end{subfigure}
\vspace{-4pt}
\caption{Fraction of ``good starts" generated during training for the robotic manipulation tasks} 
\label{fig:start_labels}
\vspace{-0.5cm}
\end{figure}

Here we evaluate if running our Procedure \ref{alg:brownian} \texttt{SampleNearby} with the start states estimated as ``good" from the previous iteration yields more good starts than running \texttt{SampleNearby} from all start states used in the previous iteration. This can clearly be seen in Figs.~\ref{fig:good_starts_key}-\ref{fig:good_starts_disc} for the robotic manipulation tasks.

\section{Conclusions and Future Directions}

We propose a method to automatically adapt the start state distribution on which an agent is trained, such that the performance on the original problem is efficiently optimized. We leverage three assumptions commonly satisfied in simulated tasks to tackle hard goal-oriented problems that state of the art RL methods cannot solve.

A limitation of the current approach is that it generates start states that grow from a single goal uniformly outwards, until they cover the original start state distribution Unif($S^0$). Nevertheless, if the target set of start states $S^0$ is far from the goal and we have some prior knowledge, it would be interesting to bias the generated start distributions $\rho_i$ towards the desired start distribution. A promising future line of work is to combine the present automatic curriculum based on start state generation with goal generation \citep{held2017goal_generation}, similar to classical results in planning \citep{kuffner2000rrt-connect}.


It can be observed in the videos of our final policy for the manipulation tasks that the agent has learned to \textit{exploit} the contacts instead of avoiding them. Therefore, the learning based aspect of the presented method has a huge potential to tackle problems that classical motion planning algorithms could struggle with, such as environments with non-rigid objects or with uncertainties in the task geometric parameters. We also leave as future work to combine our curriculum-generation approach with domain randomization methods \citep{tobin2017randomization} to obtain policies that are transferable to the real world.



\clearpage
\acknowledgments{This work was supported in part by Berkeley Deep Drive, ONR PECASE N000141612723, Darpa FunLoL. Carlos was also supported by a La Caixa Fellowship. Markus was supported by UK’s EPSRC Doctoral Training Award (DTA), and the Hans-Lenze-Foundation.}


\bibliography{start_generation}  
\clearpage

\appendix

\section{Experiment Implementation Details}

\subsection{Hyperparameters}
\label{sec:hyper}
Here we describe the hyperparemeters used for our method.  Each iteration, we generate new start states (as described in Section~\ref{sec:nearby_sampling} and Procedure~\ref{alg:brownian}), which we append to the seed states until we have a total of $M=10000$ start states.  We then subsample these down to $N_{new} = 200$ new start states.  These are appended with $N_{old}=100$ sampled old start states (as described in Section~\ref{sec:full_algo} and Procedure~\ref{alg:overall}), and these states are used to initialize our agent when we train our policy.  The ``Brownian motion" rollouts have a horizon of $T_B = 50$ timesteps, and the actions taken are random, sampled from a standard normal distribution (e.g. a 0-mean Gaussian with a covariance $\Sigma=I$).

For our method as well as the baselines, we train a $(64, 64)$ multi-layer perceptron (MLP) Gaussian policy with TRPO \citep{schulman2015trust}, implemented with rllab~\cite{duan2016benchmarking}.  We use a TRPO step-size of 0.01 and a $(32, 32)$ MLP baseline.  For all tasks, we train with a batch size of 50,000 timesteps.  All experiments use a maximum horizon of $T = 500$ time steps except for the Ant maze experiments that use a maximum horizon of $T=2000$. The episode ends as soon as the agent reaches a goal state.  We define the goal set $S^g$ to be a ball around the goal state, in which the ball has a radius of $0.03m$ for the ring and key tasks, $0.3m$ for the point-mass maze task and $0.5m$ for the ant-maze task.  In our definition of $S^0_i$, we use $R_{\rm min} = 0.1$ and $R_{\rm max} = 0.9$.  We use a discount factor $\gamma=0.998$ for the optimization, in order to encourage the policy to reach the goal as fast as possible.

\subsection{Performance metric}
\label{sec:metric}

\begin{figure}[ht]
    \captionsetup[subfigure]{justification=centering}
	\centering
	\begin{subfigure}{\linewidth}
		\centering
		\includegraphics[trim={0cm 0 0cm 0cm}, clip, width=0.16\textwidth]{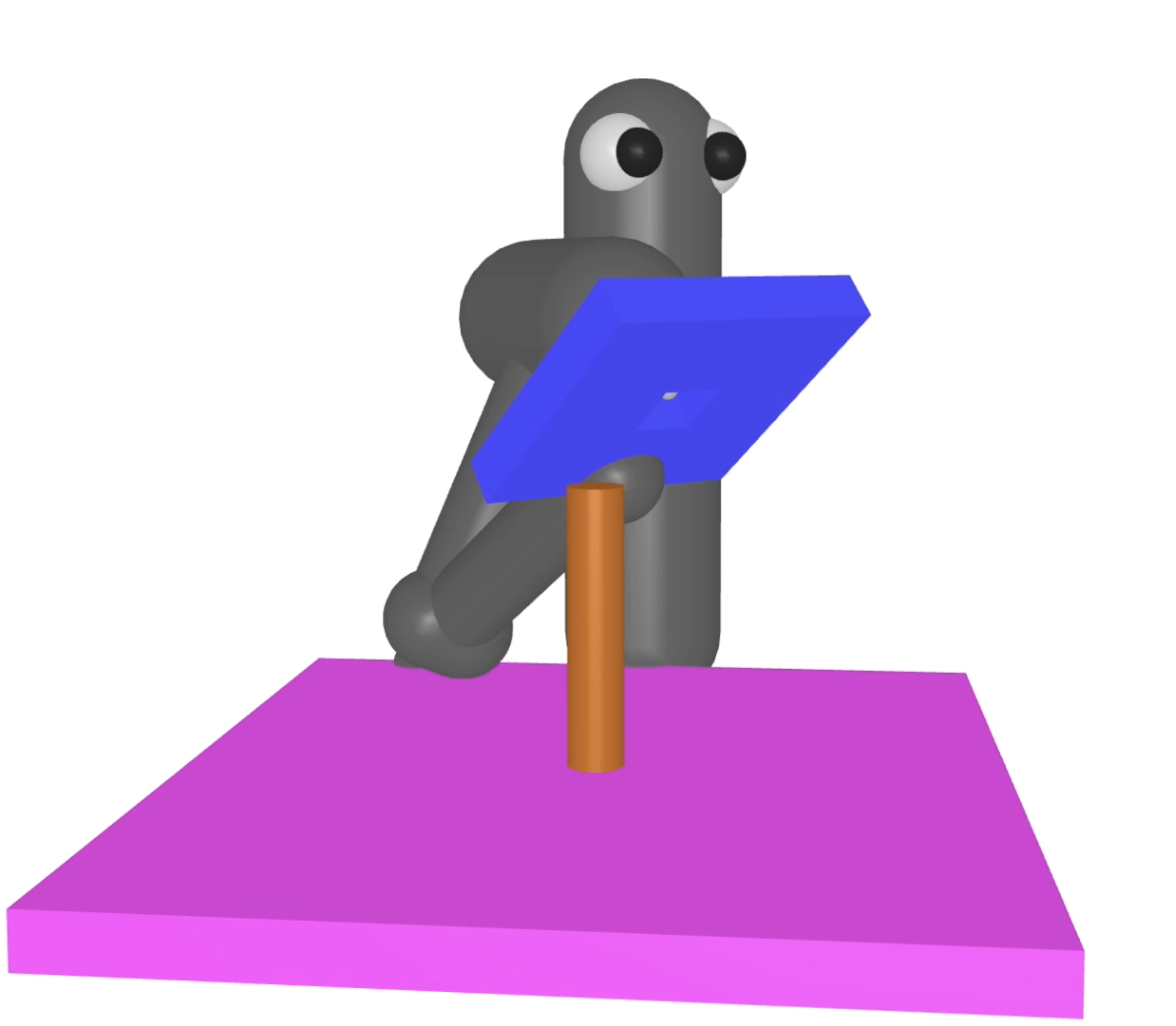}
	    \includegraphics[trim={0cm 0 0cm 0cm}, clip, width=0.16\textwidth]{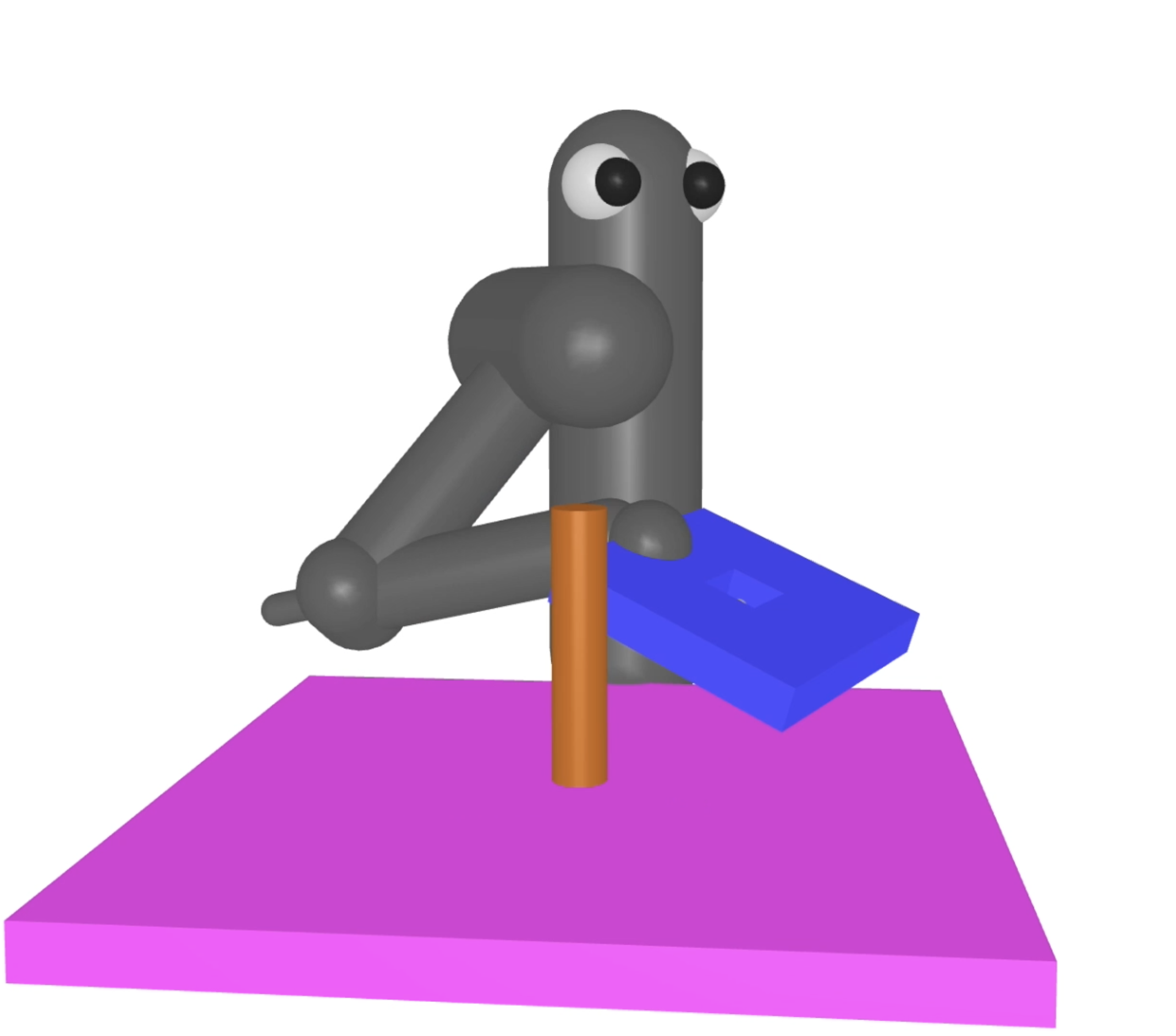}
		\includegraphics[trim={0cm 0 0cm 0cm}, clip, width=0.16\textwidth]{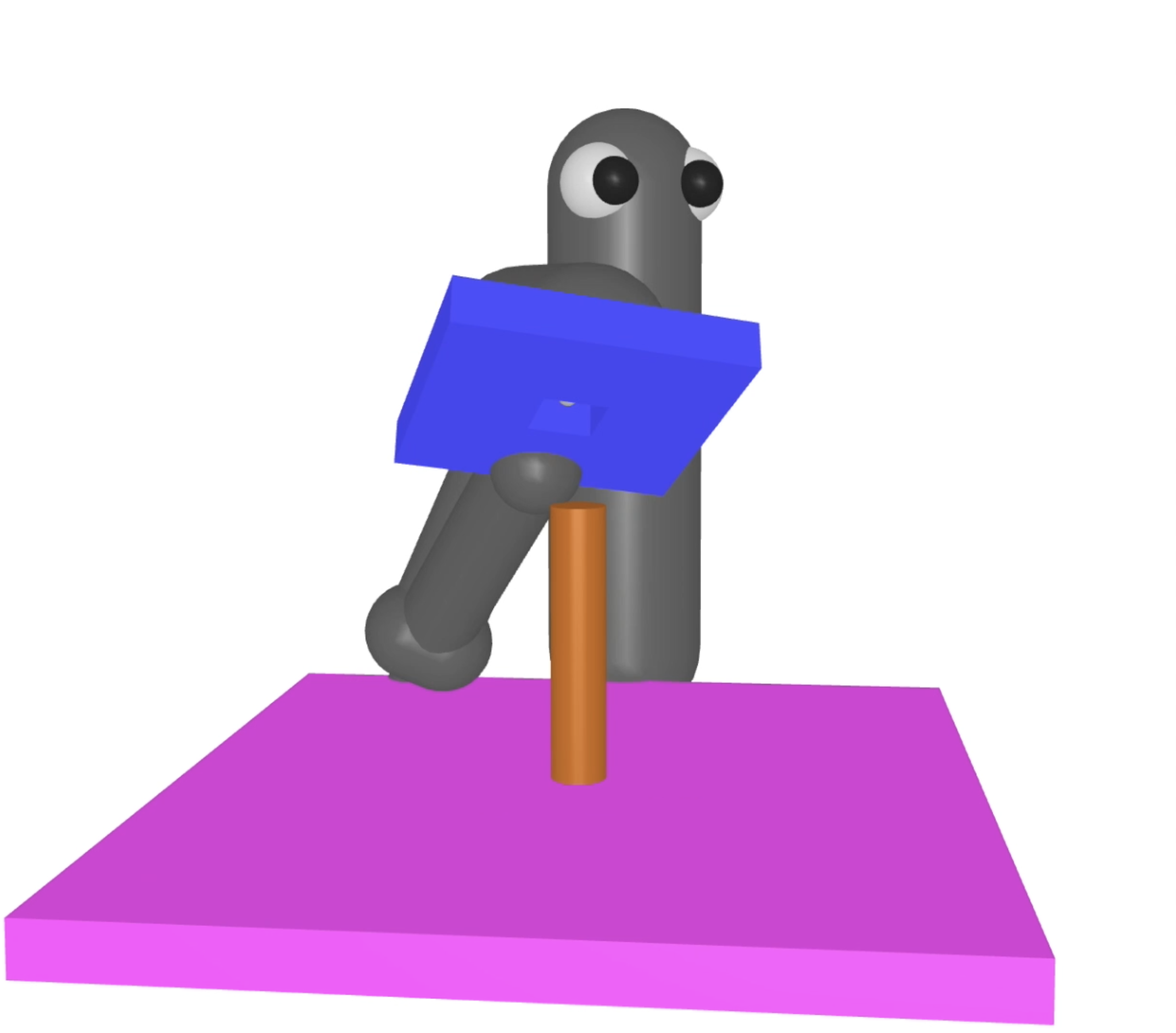}
	    \includegraphics[trim={0cm 0 0cm 0cm}, clip, width=0.16\textwidth]{Figures/disk_uniform_starts/ring_fail1.png}
		\includegraphics[trim={0cm 0 0cm 0cm}, clip, width=0.16\textwidth]{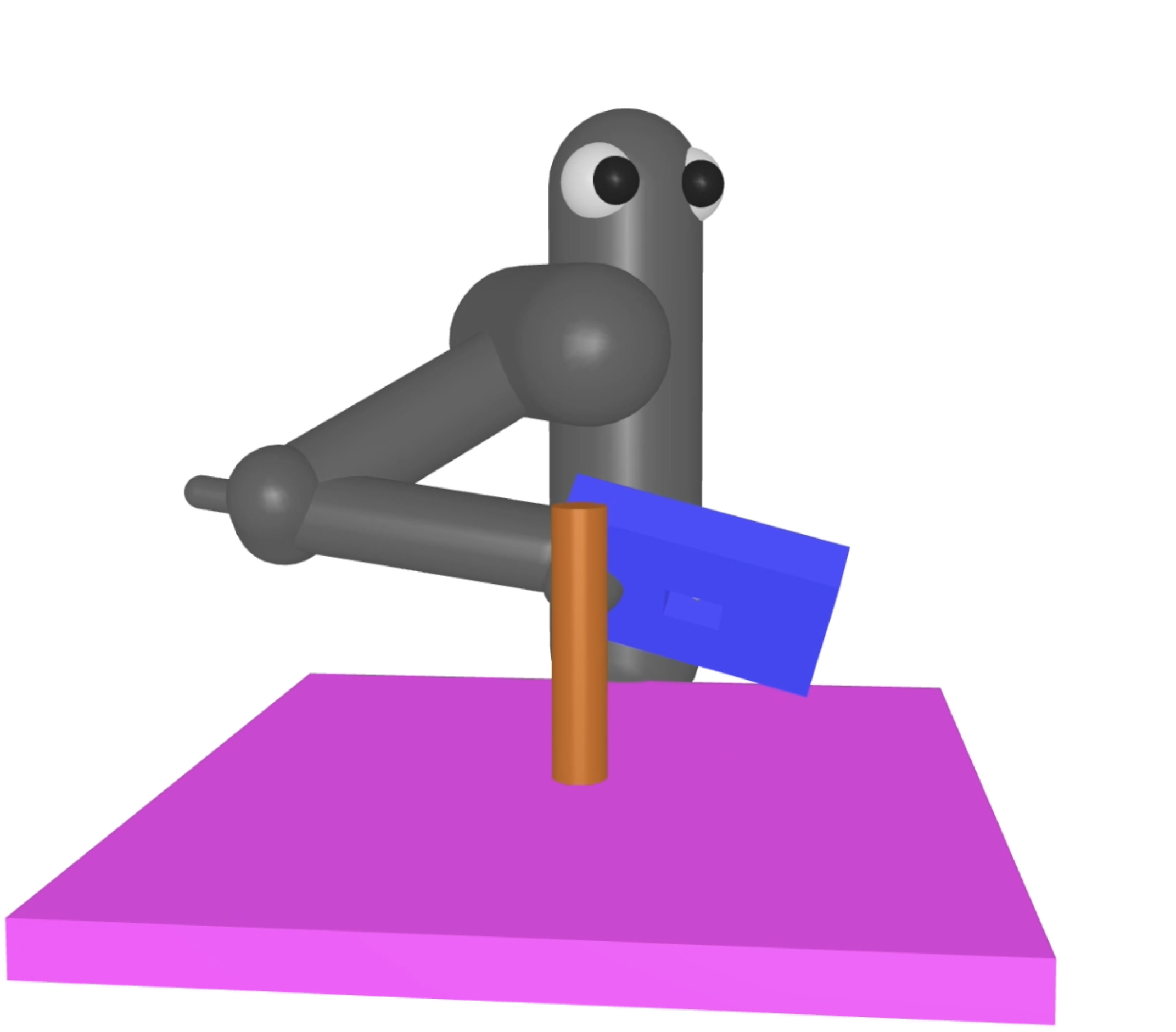}
	    \includegraphics[trim={0cm 0 0cm 0cm}, clip, width=0.16\textwidth]{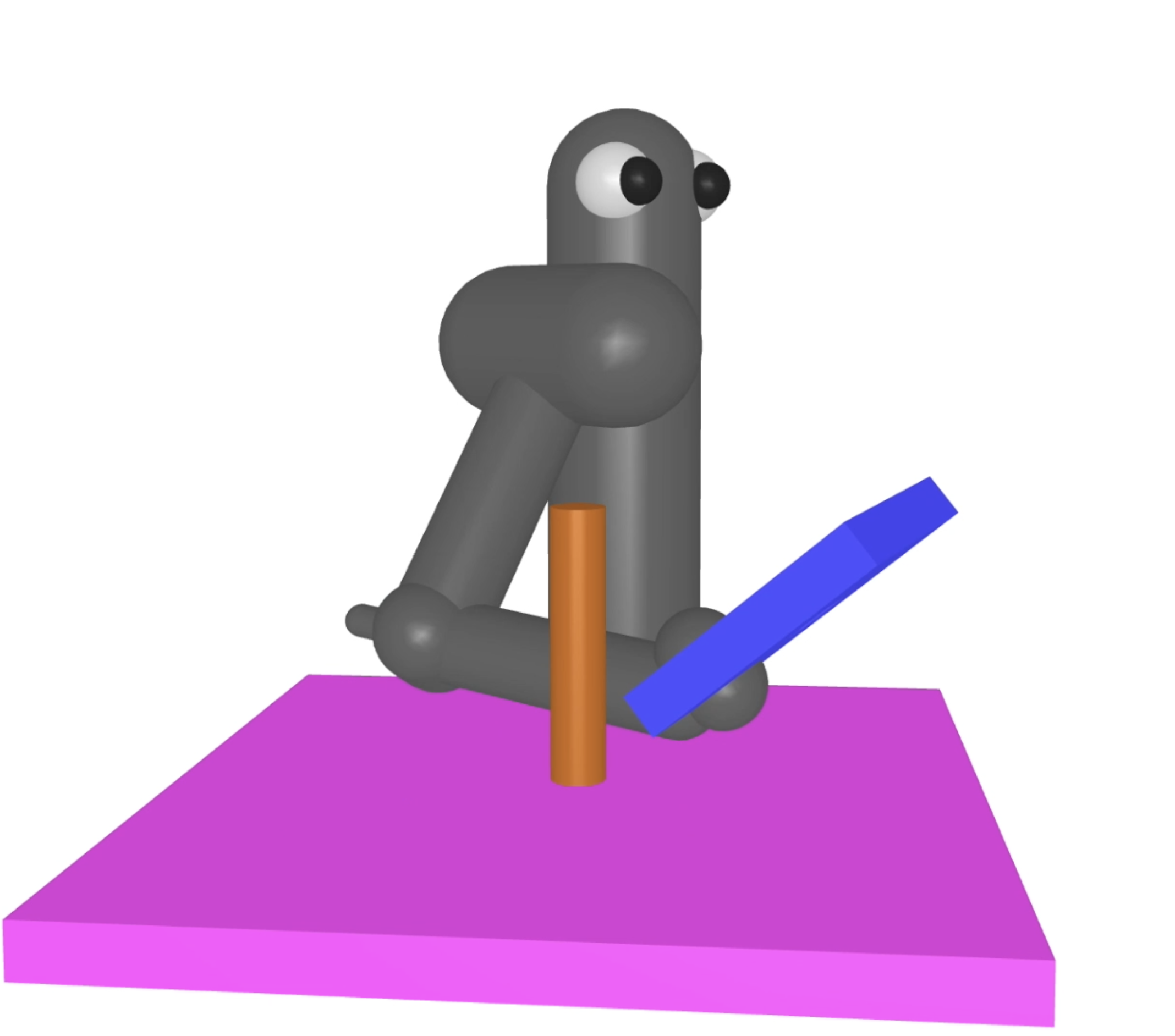}
	    \caption{Uniformly sampled start states for ring task. There are 39,530 states in the data-set, of which 5,660 have the ring with its hole already in the peg}
		\label{fig:uniform_ring}
	\end{subfigure}
	\begin{subfigure}{\linewidth}
		\centering
		\includegraphics[trim={0cm 0 0cm 0cm}, clip, width=0.16\textwidth]{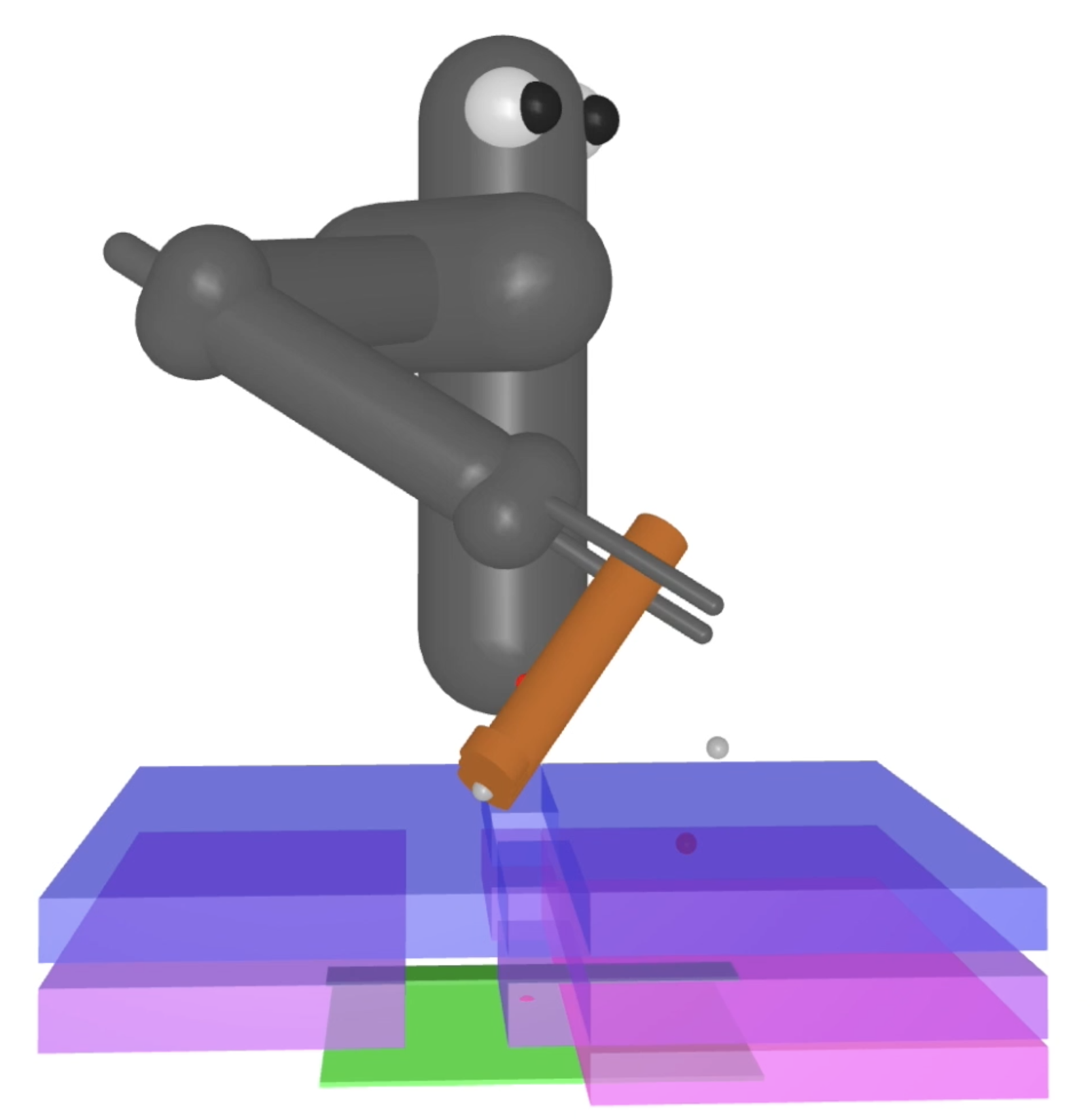}
	    \includegraphics[trim={0cm 0 0cm 0cm}, clip, width=0.16\textwidth]{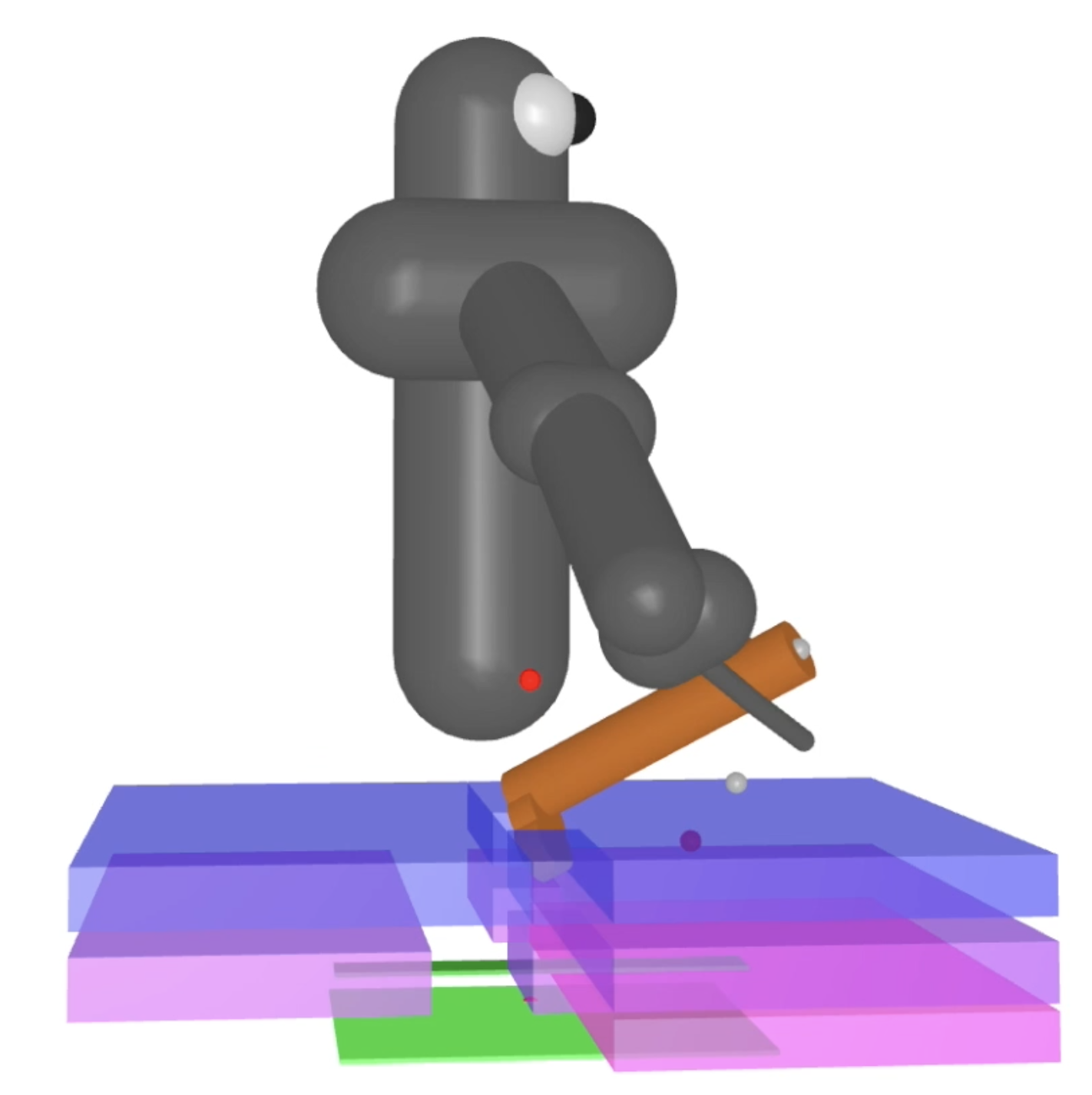}
		\includegraphics[trim={0cm 0 0cm 0cm}, clip, width=0.16\textwidth]{Figures/key_uniform_starts/key_success3.png}
	    \includegraphics[trim={0cm 0 0cm 0cm}, clip, width=0.16\textwidth]{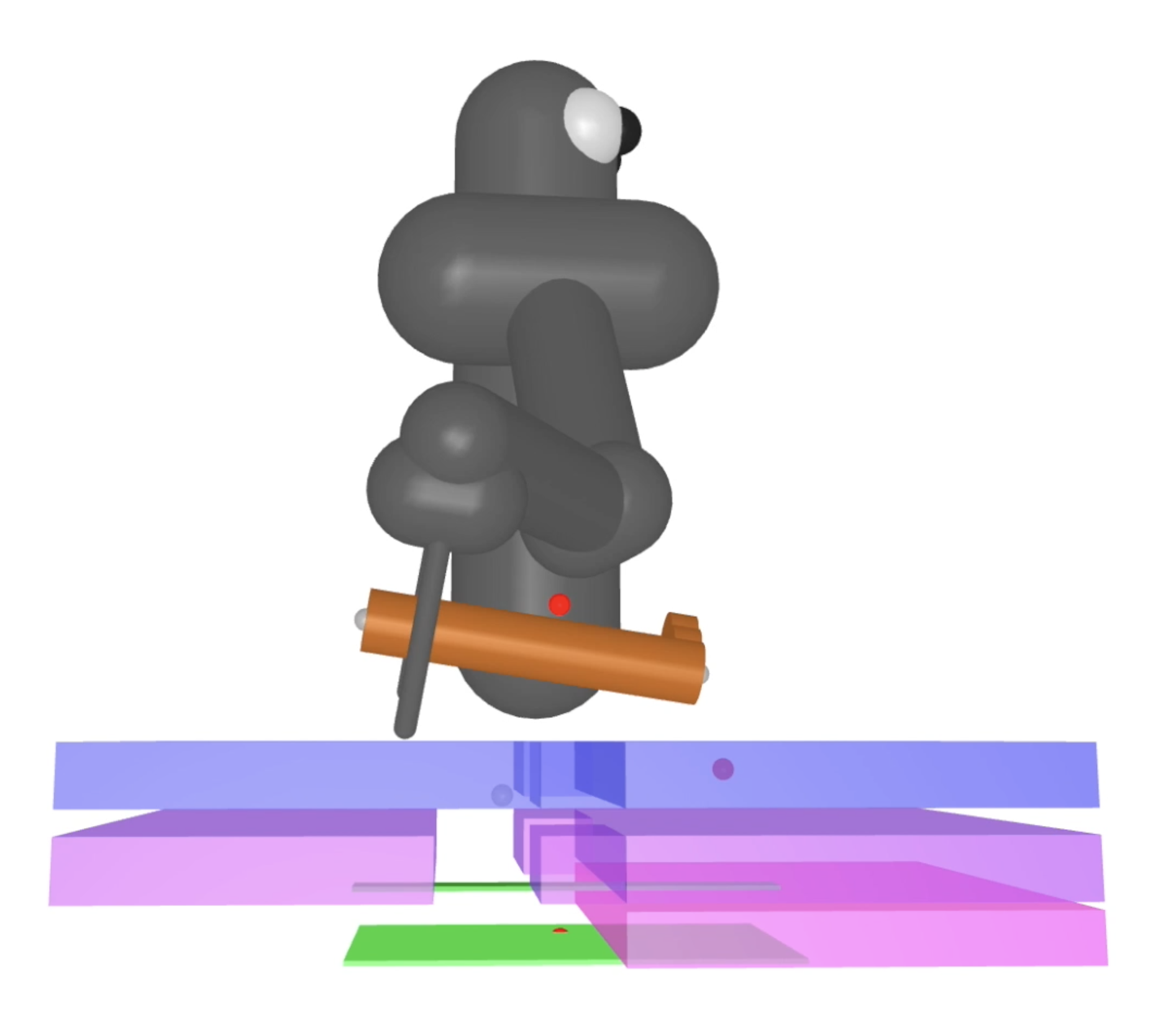}
		\includegraphics[trim={0cm 0 0cm 0cm}, clip, width=0.16\textwidth]{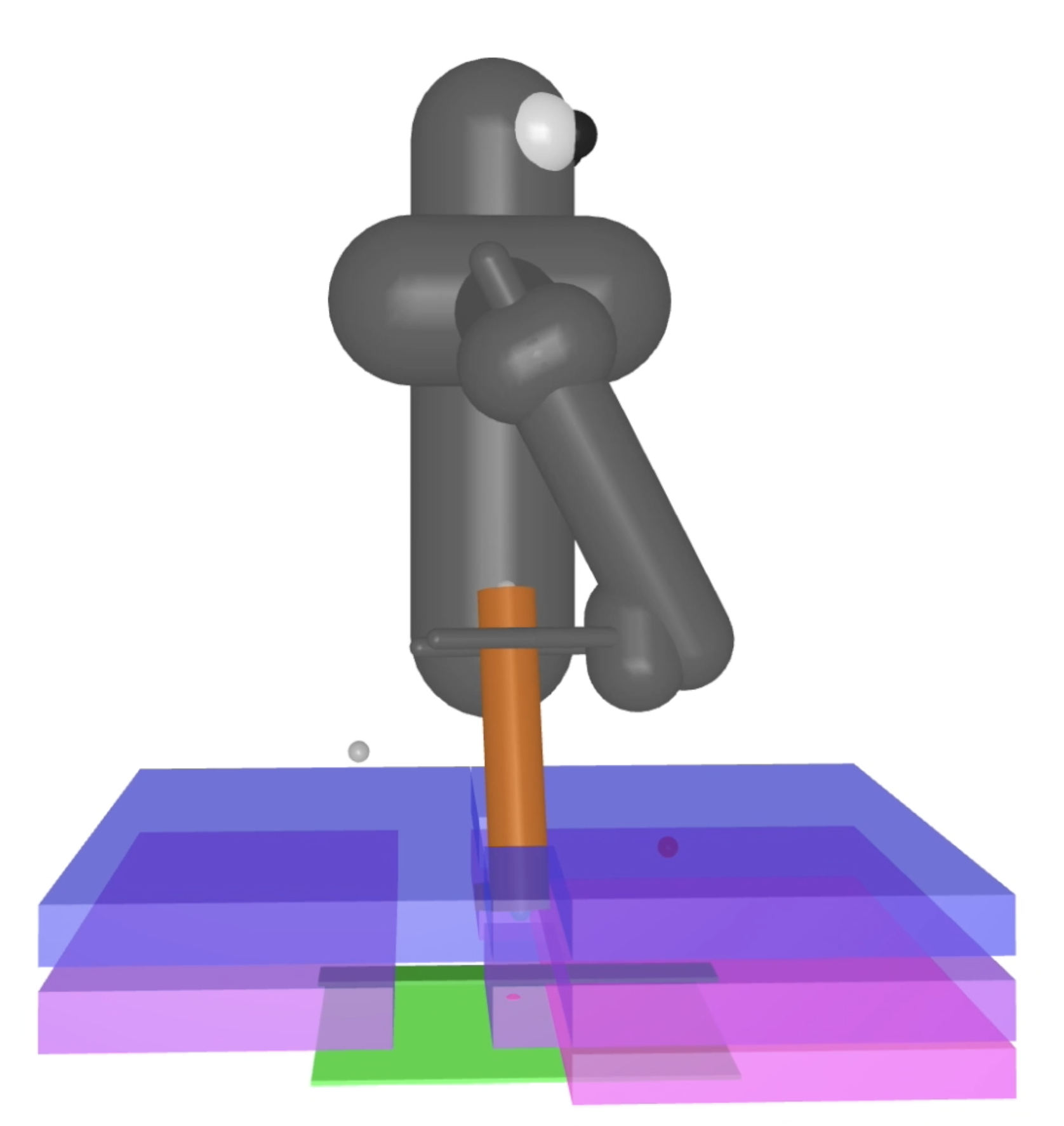}
	    \includegraphics[trim={0cm 0 0cm 0cm}, clip, width=0.16\textwidth]{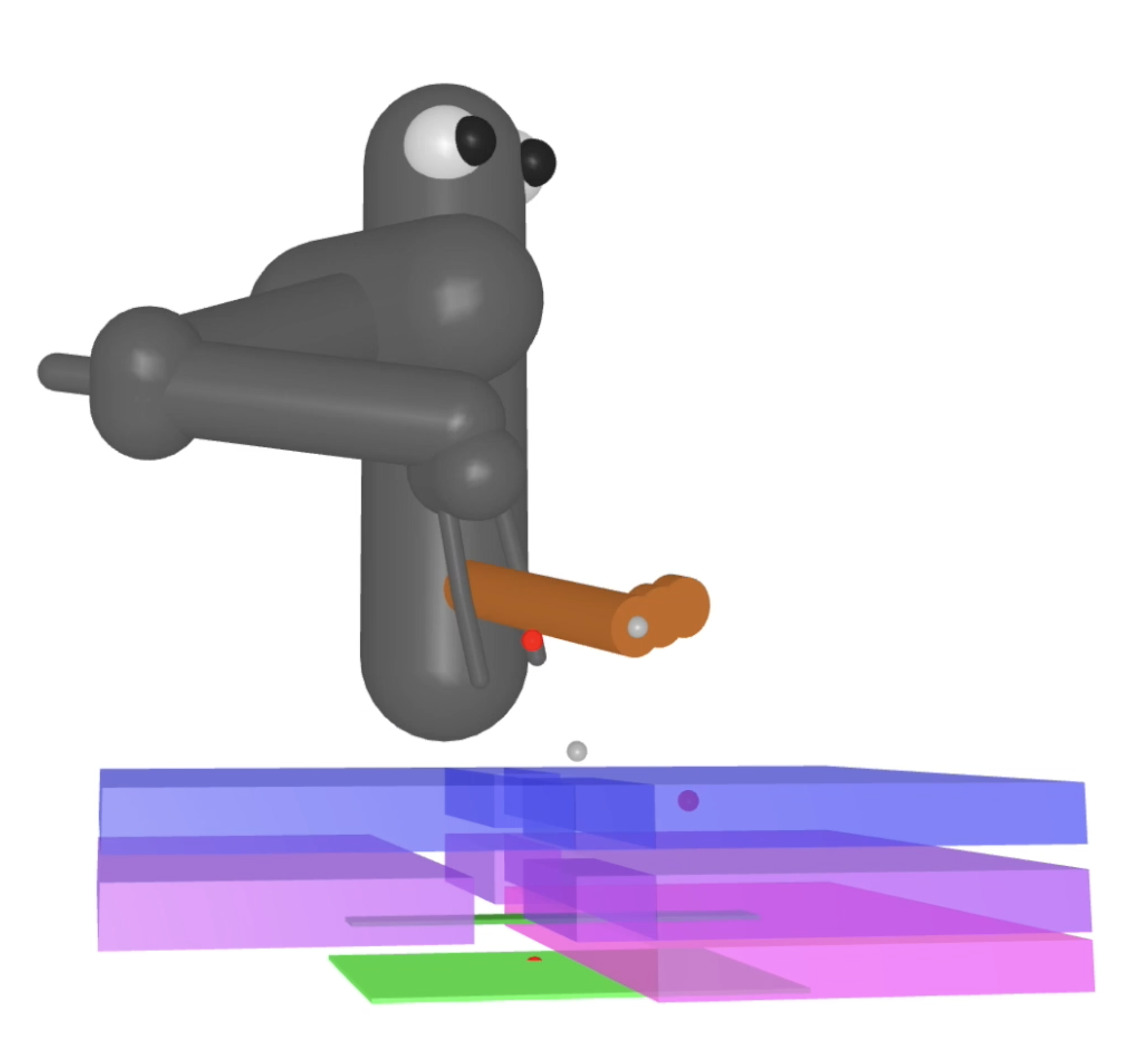}
	    \caption{Uniformly sampled start states for key task. There are 544,575 states in the data-set, of which 120,784 have the key somewhere inside the key-hole}
		\label{fig:uniform_key}
	\end{subfigure}
	\caption{Samples from the test distribution for the manipulation tasks}
	\label{fig:uniform_samples}
\end{figure}

The aim of our tasks is to reach a specified goal region $S^g$ from all start states $s_0\in S^0$ that are feasible and within a certain distance of that goal region. Therefore, to evaluate the progress on $\eta_{\rho_0}(\pi_i)$ we need to collect trajectories starting at states uniformly sampled from $S^0$. For the point-mass maze navigation task this is straight forward as the designer can give a concrete description of the feasible $(x,y)$ space, so we can uniformly sample from it. Nevertheless, it is not trivial to uniformly sample from all feasible start states for the robotics tasks. In particular, the state space is in joint angles and angular velocities of the 7 DOF arm, but the physical constraints of these contact-rich environments are given by the geometries of the task. Therefore, uniformly sampling from the angular bounds mostly yields infeasible states, with some part of the arm or the end-effector intersecting with other objects in the scene. In order to approximate uniformly sampling from $S^0$, we make use of our assumptions (Section~\ref{sec:assumptions}). We simply run our \texttt{SampleNearby} procedure initialized with $starts=[s^g]$ with a very large $M$ and long time horizons $T_B$. This large aggregated state data-set is saved and samples from it are used as proxy to $S^0$ to evaluate the performance of our algorithm. Figures \ref{fig:uniform_ring} and \ref{fig:uniform_key} show six sampled start states from the data sets used to evaluate the ring task and the key task. These data sets are available at the project website\footnote{Videos, data sets and code available at: bit.ly/reversecurriculum} for future reproducibility and benchmarking.



Given the quasi-static nature of the tasks considered, we generate only initial joint positions, and we set all initial velocities to zero. Generating initial velocities is a fairly simple extension of our approach that we leave for future work.


\section{Other methods}

\subsection{Distance reward shaping}
\begin{figure}[h]
\captionsetup[subfigure]{justification=centering}
    \centering
      \begin{subfigure}{0.48\linewidth}
        \includegraphics[width=\linewidth, trim={0cm, 0cm, 0cm, 0cm}, clip]{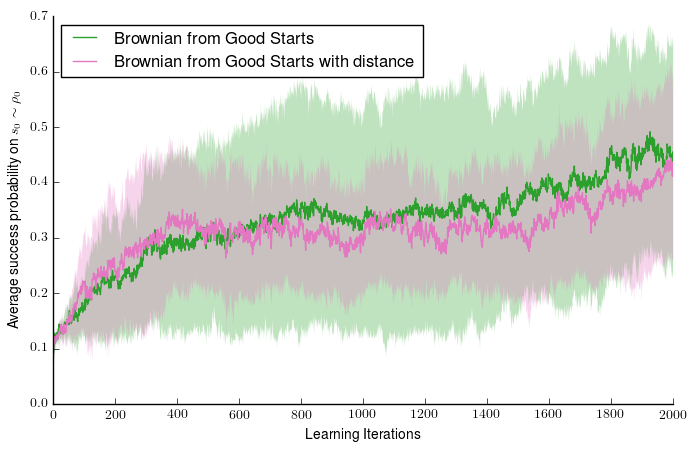}
          \caption{Ring on Peg task}
          \label{fig:learning_ring_task_dist}
      \end{subfigure}
      \begin{subfigure}{0.48\linewidth}
        \includegraphics[width=\linewidth, trim={0cm, 0cm, 0cm, 0cm}, clip]{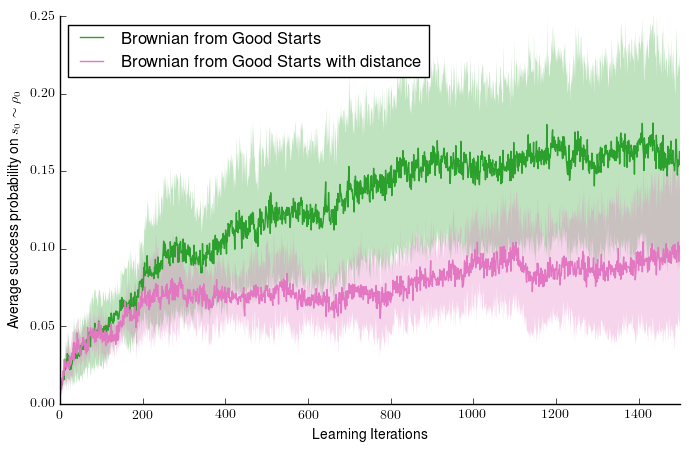}
          \caption{Key insertion task}
          \label{fig:learning_key_task_dist}
      \end{subfigure}
\caption{Learning curves for the robotics manipulation tasks} 
\label{fig:learning_dist}
\vspace{-0.2cm}
\end{figure}

Although our method is able to train policies with sparse rewards, the policy optimization steps $\texttt{train\_pol}$ can use any kind of reward shaping available. To an extent, we already do that by using a discount factor $\gamma$, which motivates the policies to reach the goal as soon as possible. Similar reward modulations could be included to take into account energy penalties or reward shaping from prior knowledge. For example, in the robotics tasks considered in this paper, the goal is defined in terms of a reference state, and hence it seems natural to try to use the distance to this state as an additional penalty to guide learning. However, we have found that this modification does not actually improve training.  For the start states near to the goal, the policy can learn to reach the goal simply from the indicator reward introduced in Section \ref{sec:task}.  For the states that are further away, the distance to the goal is actually not a useful metric to guide the policy; hence, the distance reward actually guides the policy updates towards a suboptimal local optimum, leading to poor performance. In Fig.~\ref{fig:learning_dist} we see that the ring task is not much affected by the additional reward, whereas the key task suffers considerably if this reward is added.

\subsection{Failure cases of Uniform Sampling for maze navigation}
\label{sec:failure_uniform}
In the case of the maze navigation task, we observe that applying TRPO directly on the original MDP incurs a very high variance across learning curves. We have observed that some policies only learned how to perform well from a certain side of the goal. The reason for this is that our learning algorithm (TRPO) is a batch on-policy method; therefore, at the beginning of learning, uniformly sampling from the state-space might give a batch with very few trajectories that reach the goal and hence it is likely that the successful trajectories all come from one side of the goal. In this case, the algorithm will update the policy to go in the same direction from everywhere, wrongly extrapolating from these very few successful trajectories it received. This is less likely to happen if the trajectories for the batch are collected with a different start state distribution that concentrates more uniformly around the goal, as the better learning progress of the other curves show. 

\subsection{Failure cases of Asymmetric Self-play}
\label{sec:failure_asym}
In Section~\ref{sec:results_good_starts}, we compare the performance of our method to the asymmetric self-play approach of~\citet{sukhbaatar2017intrinsic}.  Although such an approach learns faster than the uniform sampling baseline, it gets stuck in a local optimum and fails to learn to reach the goal from more than 40\% of start-states in the point-mass maze task.

As explained above, part of the reason that this method gets stuck in a local optimum is that ``Alice" (the policy that is proposing start-states) is represented with a unimodal Gaussian distribution, which is a common representation for policies in continuous action spaces.  Thus Alice's policy tends to converge to moving in a single direction.  In the original paper, this problem is somewhat mitigated by using a discrete action space, in which a multi-modal distribution for Alice can be maintained.  However, even in such a case, the authors of the original paper also observed that Alice tends to converge to a local optimum~\cite{sukhbaatar2017intrinsic}.

A further difficulty for Alice is that her reward function can be sparse, which can be inherently difficult to optimize.  Alice's reward is defined as $r_A = \max(0, t_B - t_A)$, where $t_A$ is the time that it takes Alice to reach a given start state from the goal (at which point Alice executes the ``stop" action), and $t_B$ is the time that it takes Bob to return to the goal from the start state.  Based on this reward, the optimal policy for Alice is to find the nearest state for which Bob does not know how to return to the goal; this will lead to a large value for $t_B$ with a small value for $t_A$.  In theory, this should lead to an automatic curriculum of start-states for Bob.

However, in practice, we find that sometimes, Bob's policy might improve faster than Alice's.  In such a case, Bob will have learned how to return to the goal from many start states much faster than Alice can reach those start states from the goal.  In such cases, we would have that $t_B < t_A$, and hence $r_A = 0$.  Thus, Alice's rewards are sparse (many actions that Alice takes result in 0 reward) and hence it will be difficult for Alice's policy to improve, leading to a locally optimal policy for Alice.  For these reasons, we have observed Alice's policy often getting ``stuck," in which Alice is unable to find new start-states to propose for Bob that Bob does not already know how to reach the goal from.

We have implemented a simple environment that illustrates these issues.  In this environment, we use a synthetic ``Bob" that can reach the goal from any state within a radius $r_B$ from the goal.  For states within $r_B$, Bob can reach the goal in a time proportional to the distance between the state and the goal; in other words, for such states $s_0 \in \{s : |s - s^g| < r_B, s \in S^0 \}$, we have that $t_B = |s_0 - s^g| / v_B$, where $|s_0 - s^g|$ is the distance between state $s_0$ and the goal $s^g$, and $v_B$ is Bob's speed.  For states further than $r_B$ from the goal, Bob does not know how to reach the goal, and thus $t_B$ for such states takes the maximum possible value.

\begin{figure}[ht]
\captionsetup[subfigure]{justification=centering}
    \centering
      \begin{subfigure}{0.3\linewidth}
        \includegraphics[width=\linewidth, trim={0cm, 0cm, 0cm, 0cm}, clip]{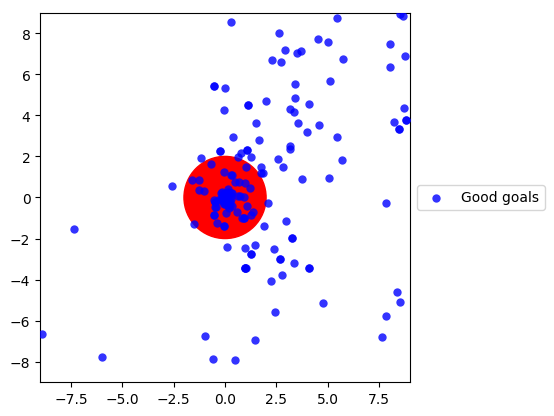}
          \caption{Iteration 1}
          \label{fig:alice1}
      \end{subfigure}
      \begin{subfigure}{0.33\linewidth}
        \includegraphics[width=\linewidth, trim={0cm, 0cm, 0cm, 0cm}, clip]{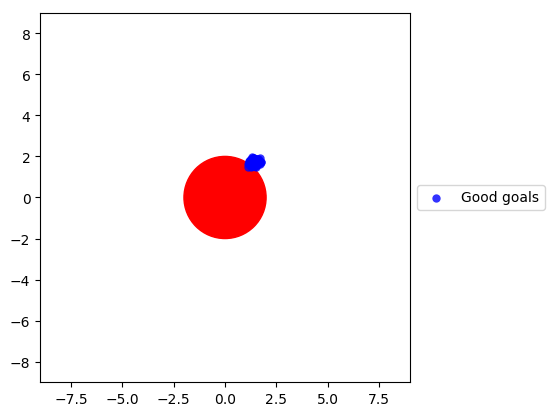}
          \caption{Iteration 10}
          \label{fig:alice10}
      \end{subfigure}
      \begin{subfigure}{0.33\linewidth}
        \includegraphics[width=\linewidth, trim={0cm, 0cm, 0cm, 0cm}, clip]{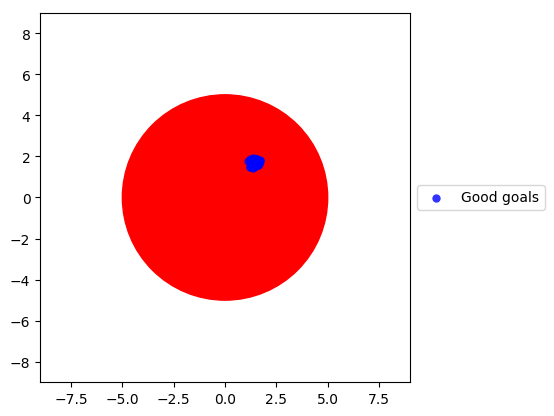}
          \caption{Iteration 32}
          \label{fig:alice32}
      \end{subfigure}
\caption{Simple environment to illustrate asymmetric self-play~\cite{sukhbaatar2017intrinsic}.  The red areas indicate the states from which Bob knows how to reach the goal.  The blue points are the start-states proposed by Alice at each iteration (i.e. the states from which Alice performed the stop action)}.
\label{fig:asym_simple}
\vspace{-0.2cm}
\end{figure}

This setup is illustrated in Figure~\ref{fig:asym_simple}.  The region shown in red designates the area within $r_B$ from the goal, e.g. the set of states from which Bob knows how to reach the goal.  On the first iteration, Alice has a random policy (Figure~\ref{fig:alice1}).  After 10 iterations of training, Alice has converged to a policy that reaches the location just outside of the set of states from which Bob knows how to reach the goal (Figure~\ref{fig:alice10}).  From these states, Alice receives a maximum reward, because $t_B$ is very large while $t_A$ is low.  Note that we also observe the unimodal nature of Alice's policy; Alice has converged to a policy which proposes just one small set of states among all possible states for which she would receive a similar reward.

At this point we synthetically increase $r_B$, corresponding to the situation in which Bob learns how to reach the goal from a larger set of states.  However, Alice's policy has already converged to reaching a small set of states which were optimal for Bob's previous policy.  From these states Alice now receives a reward of 0, as described above: Bob can return from these states quickly to the goal, so we have that $t_B < t_A$ and $r_A = 0$.  Thus, Alice does not receive any reward signal and is not able to improve her policy.  Hence, Alice's policy remains stuck at this point and she is not able to find new states to propose to Bob (Figure~\ref{fig:alice32}).  

In this simple case, one could attempt to perform various hacks to try to fix the situation, e.g. by artificially increasing Alice's variance, or by resetting Alice to a random policy.  However, note that, in a real example, Bob is learning an increasingly complex policy, and so Alice would need to learn an equally complex policy to find a set of states that Bob cannot succeed from; hence, these simple fixes would not suffice to overcome this problem.  Fundamentally, the asymmetric nature of the self-play between Alice and Bob creates a situation in which Alice has a difficult time learning and often gets stuck in a local optimum from which she is unable to improve.


\end{document}